\documentclass[letterpaper]{article} 
\usepackage{aaai23}  
\usepackage{times}  
\usepackage{helvet}  
\usepackage{courier}  
\usepackage[hyphens]{url}  
\usepackage{graphicx} 
\usepackage{natbib}  
\usepackage{caption} 
\usepackage{algorithm}

\usepackage{newfloat}
\usepackage{listings}
\DeclareCaptionStyle{ruled}{labelfont=normalfont,labelsep=colon,strut=off} 
\floatstyle{ruled}
\newfloat{listing}{tb}{lst}{}
\floatname{listing}{Listing}
\title{FedALA: Adaptive Local Aggregation for Personalized Federated Learning}
\author {
    Jianqing Zhang\textsuperscript{\rm 1},
    Yang Hua\textsuperscript{\rm 2},
    Hao Wang\textsuperscript{\rm 3},
    Tao Song\textsuperscript{\rm 1},
    Zhengui Xue\textsuperscript{\rm 1},
    Ruhui Ma\textsuperscript{\rm 1}\thanks{Corresponding author.},
    Haibing Guan\textsuperscript{\rm 1}
}
\affiliations {
    \textsuperscript{\rm 1}Shanghai Jiao Tong University \\
    \textsuperscript{\rm 2}Queen's University Belfast \\
    \textsuperscript{\rm 3}Louisiana State University \\
    \{tsingz, songt333, zhenguixue, ruhuima, hbguan\}@sjtu.edu.cn, Y.Hua@qub.ac.uk, haowang@lsu.edu
}

\usepackage{tikz}
\usepackage{comment}
\usepackage{amsmath,amssymb} 

\usepackage{graphicx}
\usepackage{booktabs}

\DeclareMathOperator*{\argmin}{arg\,min}
\usepackage{mathtools}
\usepackage{multirow}
\usepackage[noend]{algpseudocode}
\usepackage{subfigure}

\usepackage{xspace}
\def\method{FedALA\xspace}
\def\module{ALA\xspace}

\usepackage{cleveref}

\usepackage{cite}

\makeatletter
\DeclareRobustCommand\onedot{\futurelet\@let@token\@onedot}
\def\@onedot{\ifx\@let@token.\else.\null\fi\xspace}
\def\eg{\emph{e.g}\onedot} 
\def\ie{\emph{i.e}\onedot}

\makeatother

\usepackage{multicol}
\algnewcommand{\LineComment}[1]{\State \(\triangleright\) #1}
\makeatletter
\let\OldStatex\Statex
\renewcommand{\Statex}[1][3]{%
  \setlength\@tempdima{\algorithmicindent}%
  \OldStatex\hskip\dimexpr#1\@tempdima\relax}
\makeatother

\definecolor{blue_}{RGB}{76, 114, 176}
\definecolor{orange_}{RGB}{221, 132, 82}
\definecolor{upload}{RGB}{47, 85, 151}
\definecolor{download}{RGB}{241, 13, 208}
\definecolor{red_}{RGB}{255, 0, 0}
\definecolor{gray_}{RGB}{127, 127, 127}
\definecolor{green_}{RGB}{1, 128, 0}

\begin{document}

\maketitle

\begin{abstract}

A key challenge in federated learning (FL) is the statistical heterogeneity that impairs the generalization of the global model on each client. To address this, we propose a method \emph{\textbf{Fed}erated learning with \textbf{A}daptive \textbf{L}ocal \textbf{A}ggregation} (\textbf{\method}) by capturing the desired information in the global model for client models in personalized FL. The key component of \method is an \emph{\textbf{A}daptive \textbf{L}ocal \textbf{A}ggregation} (\textbf{\module}) module, which can adaptively aggregate the downloaded global model and local model towards the local objective on each client to initialize the local model before training in each iteration. To evaluate the effectiveness of \method, we conduct extensive experiments with five benchmark datasets in computer vision and natural language processing domains. \method outperforms eleven state-of-the-art baselines by up to 3.27\% in test accuracy. Furthermore, we also apply \module module to other federated learning methods and achieve up to 24.19\% improvement in test accuracy. Code is available at https://github.com/TsingZ0/FedALA.

\end{abstract}

\section{Introduction}
\label{sec:intro}

Federated learning (FL) can leverage distributed user data while preserving privacy by iteratively downloading models, training models locally on the clients, uploading models, and aggregating models on the server. A key challenge in FL is statistical heterogeneity, \eg, the not independent and identically distributed (Non-IID) and unbalanced data across clients. This kind of data makes it hard to obtain a global model that generalizes to each client~\cite{mcmahan2017communication, reisizadeh2020fedpaq, t2020personalized}. 

Personalized FL (pFL) methods have been proposed to tackle statistical heterogeneity in FL. Unlike traditional FL that seeks a high-quality global model via distributed training across clients, \eg, FedAvg~\cite{mcmahan2017communication}, pFL methods are proposed to prioritize the training of a local model for each client. Recent pFL studies on model aggregation on the server can be classified into three categories: (1) methods that learn a single global model and fine-tune it, including Per-FedAvg~\cite{NEURIPS2020_24389bfe} and FedRep~\cite{collins2021exploiting}, (2) methods that learn additional personalized models, including pFedMe~\cite{t2020personalized} and Ditto~\cite{li2021ditto}, and (3) methods that learn local models with personalized (local) aggregation, including FedAMP~\cite{huang2021personalized}, FedPHP~\cite{li2021fedphp}, FedFomo~\cite{zhang2020personalized}, APPLE~\cite{luo2021adapt} and PartialFed~\cite{sun2021partialfed}. 

pFL methods in Category (1) and (2) take all the information in the global model for local initialization, \ie, initializing the local model before local training in each iteration. However, only the desired information that improves the quality of the local model is beneficial for the client. The global model has poor generalization ability since it has desired and undesired information for an individual client simultaneously. Thus, pFL methods in Category (3) intend to capture the desired information in the global model through personalized aggregation. 

However, pFL methods in Category (3) still have shortcomings. FedAMP/FedPHP performs personalized aggregation on the server/clients without considering the local objective. FedFomo/APPLE downloads other client models and locally aggregates them with the approximated/learned aggregating weights on each client. All the parameters in one client model are assigned with the same weight, \ie, model-level weight. Besides, downloading client models among clients causes high communication overhead in each iteration and also has privacy concerns since the data from other clients can be recovered through these client models~\cite{zhu2019deep}. In addition, FedFomo/APPLE also requires feeding data forward in the downloaded client models to obtain the aggregating weights, which introduces additional computation overhead. PartialFed locally learns aggregation strategies to select the parameters in the global model or the local model. Still, the layer-level and binary selection cannot precisely capture the desired information in the global model. Furthermore, PartialFed uses non-overlapping samples to learn the local model and the strategy, so it can hardly learn a strategy that fully satisfies the local objective. Due to the significant modification of the learning process in FedAvg, the personalized aggregation process in these methods cannot be directly applied to most existing FL methods. 

\begin{figure}[t]
	\centering
	\includegraphics[width=\linewidth]{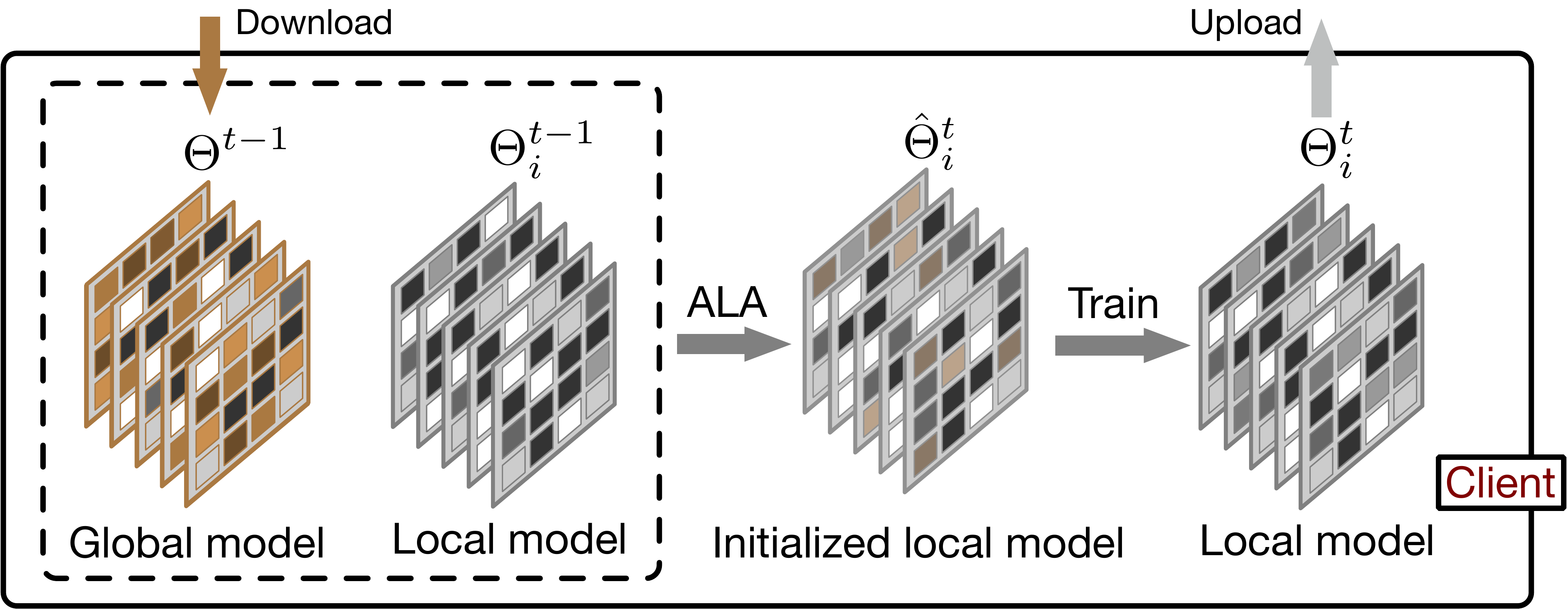}
	\caption{Local learning process on client $i$ in the $t$-th iteration. Specifically, client $i$ downloads the global model from the server, locally aggregates it with the old local model by \module module for local initialization, trains the local model, and finally uploads the trained local model to the server.}
	\label{fig:illustrate}
\end{figure}

To precisely capture the desired information in the downloaded global model for each client without additional communication overhead in each iteration, we propose a novel pFL method \emph{\textbf{Fed}erated learning with \textbf{A}daptive \textbf{L}ocal \textbf{A}ggregation} (\textbf{\method}) to adaptively aggregates the downloaded global model and local model towards the local objective for local initialization. \method only downloads one global model and uploads one local model on each client with the same communication overhead as in FedAvg, which also has fewer privacy concerns and is more communication-efficient than FedFomo and APPLE. By adaptively learning real-value and element-wise aggregation weights towards the local objective on the full local dataset, \method can capture the desired information in the global model at the element level, which is more precise than the binary and layer-wise weight learning in PartialFed. 
Since the lower layers in a deep neural network (DNN) learn more generic information than the higher layers~\cite{yosinski2014transferable, lecun2015deep}, we can further reduce the computation overhead by only applying \emph{\textbf{A}daptive \textbf{L}ocal \textbf{A}ggregation} (\textbf{\module}) module on the higher layers. The whole local learning process is shown in \Cref{fig:illustrate}. 

To evaluate the effectiveness of \method, we conduct extensive experiments on five benchmark datasets. 
Results show that \method outperforms eleven state-of-the-art (SOTA) methods. 
Besides, we apply the \module module to the traditional FL methods and pFL methods to improve their performance. 
To sum up, our contributions are as follows.
\begin{itemize}
    \item We propose a novel pFL method \method that adaptively aggregates the global model and local model towards the local objective to capture the desired information from the global model in an element-wise manner. 
    \item We empirically show the effectiveness of \method, which outperforms eleven SOTA methods by up to 3.27\% in test accuracy without additional communication overhead in each iteration. 
    \item Attributed to the minor modification of FL process, the \module module in \method can be directly applied to existing FL methods to enhance their performance by up to 24.19\% in test accuracy on Cifar100. 
\end{itemize}

\section{Related Work}

\subsection{Traditional Federated Learning}

The widely known traditional FL method FedAvg~\cite{mcmahan2017communication} learns a single global model for all clients by aggregating their local models. However, it often suffers in statistically heterogeneous settings, \eg, FL with Non-IID and unbalanced data~\cite{kairouz2019advances, zhao2018federated}. To address this issue, FedProx~\cite{MLSYS2020_38af8613} improves the stability of the FL process through a proximal term. To counteract the bias introduced by the Non-IID data, FAVOR~\cite{wang2020optimizing} selects a subset of clients based on deep Q-learning~\cite{van2016deep} at each iteration. By generating the global model layer-wise, FedMA~\cite{wang2020federated} can adapt to statistical heterogeneity with the matched averaging approach. However, with statistical heterogeneity in FL, it is hard to obtain a single global model which generalizes well to each client~\cite{kairouz2019advances, huang2021personalized, t2020personalized}.

\subsection{Personalized Federated Learning}

Recently, personalization has attracted much attention for tackling statistical heterogeneity in FL~\cite{kairouz2019advances}. We consider the following three categories of pFL methods that focus on aggregating models on the server: 

\noindent\textbf{(1) Methods that learn a single global model and fine-tune it.} Per-FedAvg~\cite{NEURIPS2020_24389bfe} considers the global model as an initial shared model based on MAML~\cite{finn2017model}. By performing a few additional training steps locally, all the clients can easily fine-tune the reasonable initial shared model. FedRep~\cite{collins2021exploiting} splits the backbone into a global model (representation) and a client-specific head and fine-tunes the head locally to achieve personalization. 

\noindent\textbf{(2) Methods that learn additional personalized models.} pFedMe\cite{t2020personalized} learns an additional personalized model for each client with Moreau envelopes. In Ditto~\cite{li2021ditto}, each client learns its additional personalized model with a proximal term to fetch information from the downloaded global model. 

\noindent\textbf{(3) Methods that learn local models with personalized (local) aggregation.} To further capture the personalization, recent methods try to generate client-specific models through personalized aggregation. For example, FedAMP~\cite{huang2021personalized} generates an aggregated model for an individual client by the attention-inducing function and personalized aggregation. To utilize the historical local model, FedPHP~\cite{li2021fedphp} locally aggregates the global model and local model with the rule-based moving average and a pre-defined weight (hyperparameter). FedFomo~\cite{zhang2020personalized} focuses on aggregating other client models locally for local initialization in each iteration and approximates the client-specific weights for aggregation using the local models from other clients. Similar to FedFomo, APPLE~\cite{luo2021adapt} also aggregates client models locally but learns the weights instead of approximation and performs the local aggregation in each training batch rather than just local initialization. FedAMP and APPLE are proposed for the cross-silo FL setting, which require all clients to join each iteration. By learning the binary and layer-level aggregation strategy for each client, PartialFed~\cite{sun2021partialfed} selects the parameters in the global model or the local model to construct the new local model in each batch. 

Our \method belongs to Category (3) but is more precise and requires less communication cost than FedFomo and APPLE. The fine-grained \module in \method can element-wisely aggregate the global model and local model to adapt to the local objective on each client for local initialization. As \method only modifies the local initialization in FL, it can be applied to existing FL methods to improve their performance without modifying other learning processes.

\section{Method}

\subsection{Problem Statement}

Suppose we have $N$ clients with their private training data $D_1, \ldots, D_N$, respectively. These datasets are heterogeneous (Non-IID and unbalanced). Specifically, $D_1, \ldots, D_N$ are sampled from $N$ distinct distributions and have different sizes. With the help of a central server, our goal is to collaboratively learn individual local models $\hat{\Theta}_1, \ldots, \hat{\Theta}_N$ using $D_1, \ldots, D_N$ for each client, without exchanging the private data. 
The objective is to minimize the global objective and obtain the reasonable local models
\begin{equation}
    \{\hat{\Theta}_1, \ldots, \hat{\Theta}_N\} = \argmin \ \mathcal{G}(\mathcal{L}_1, \ldots, \mathcal{L}_N),
\end{equation}
where $\mathcal{L}_i = \mathcal{L}(\hat{\Theta}_i, D_i; \Theta), \forall i\in [N]$ and $\mathcal{L}(\cdot)$ is the loss function. $\Theta$ is the global model, which brings external information to client $i$. Typically, $\mathcal{G}(\cdot)$ is set to $\sum^{N}_{i=1} k_i \mathcal{L}_i$, where $k_i = |D_i| / \sum^{N}_{j=1} |D_j|$ and $|D_i|$ is the amount of local data samples on client $i$. 

\subsection{Adaptive Local Aggregation (\module)}

\begin{figure}[ht]
	\centering
	\includegraphics[width=\linewidth]{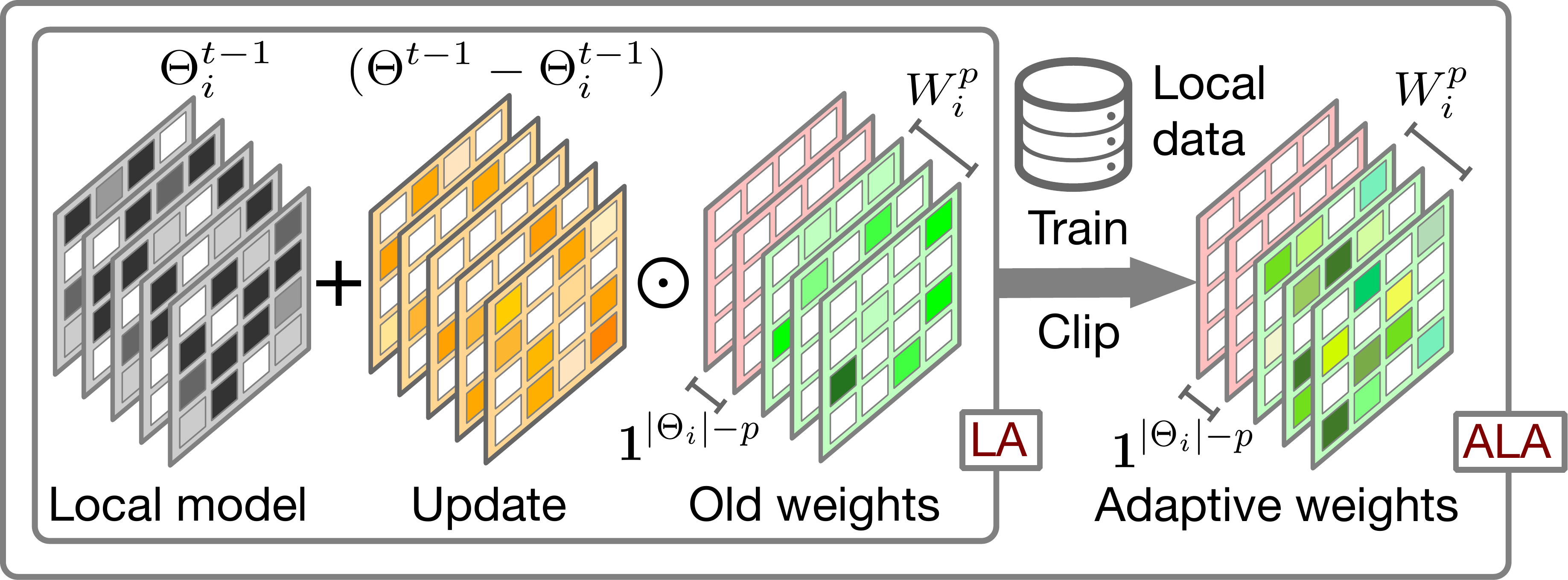}
	\caption{The learning process in \module. LA denotes ``local aggregation''. Here, we consider a five-layer model and set $p=3$. The lighter the color, the larger the value. }
	\label{fig:main}
\end{figure}

The server generates a global model by aggregating trained client models in heterogeneous settings, but this global model generalizes poorly on each client. To address this problem, we propose a pFL method \method with a fine-grained \module module that element-wisely aggregates the global model and local model to adapt to the local objective. We show the learning process in \module in \Cref{fig:main}. 

In traditional FL (\eg, FedAvg), after the server sends the old global model $\Theta^{t-1}$ to client $i$ in iteration $t$, $\Theta^{t-1}$ overwrites the old local model $\Theta^{t-1}_i$ to obtain the initialized local model $\hat{\Theta}^{t}_i$ for local model training, \ie,  $\hat{\Theta}^{t}_i := \Theta^{t-1}$. In \method, we element-wisely aggregate the global model and local model instead of overwriting. Formally, 
\begin{equation}
\begin{aligned}
    \hat{\Theta}^{t}_i &:= \Theta^{t-1}_i \odot W_{i, 1} + \Theta^{t-1} \odot W_{i, 2}, \\
    & s.t.\quad w^q_1 + w^q_2 = 1, \forall \ {\rm valid} \ q \label{eq:two_w}
\end{aligned}
\end{equation}
where $\odot$ is a Hadamard product and $w^q_1$ and $w^q_2$ are the $q$-th parameters in the aggregating weights $W_{i, 1}$ and $W_{i, 2}$, respectively. For overwriting, 
$\forall \ {\rm valid} \ q, w^q_1 \equiv 0$ and $w^q_2 \equiv 1$. 

However, it is hard to learn $W_{i, 1}$ and $W_{i, 2}$ with the constraint through the gradient-based learning method. Thus, we combine $W_{i, 1}$ and $W_{i, 2}$ by viewing \Cref{eq:two_w} as:
\begin{equation}
    \hat{\Theta}^{t}_i := \Theta^{t-1}_i + (\Theta^{t-1} - \Theta^{t-1}_i) \odot W_i,
\end{equation}
where we call the term $(\Theta^{t-1} - \Theta^{t-1}_i)$ as the ``update''. Inspired by the previous methods~\cite{courbariaux2016binarized, luo2018adaptive}, we utilize element-wise weight clipping $\sigma(w) = \max(0, \min(1, w))$ for regularization~\cite{arjovsky2017wasserstein} and let $w \in [0, 1], \forall w \in W_i$. 

Since the lower layers in the DNN learn more general information than the higher layers~\cite{yosinski2014transferable, pmlr-v139-zhu21b}, the client desires most of the information in the lower layers of the global model. To reduce computation overhead, we introduce a hyperparameter $p$ to control the range of \module by applying it on $p$ higher layers and overwriting the parameters in the lower layers like FedAvg for local initialization: 
\begin{equation}
    \hat{\Theta}^{t}_i := \Theta^{t-1}_i + (\Theta^{t-1} - \Theta^{t-1}_i) \odot [\textbf{1}^{|\Theta_i|-p}; W^p_i], \label{eq:extract-}
\end{equation}
where $|\Theta_i|$ is the number of layers (or blocks) in $\Theta^{t-1}_i$ and $\textbf{1}^{|\Theta_i|-p}$ has the same shape of the lower layers in $\Theta^{t-1}_i$. The elements in $\textbf{1}^{|\Theta_i|-p}$ are ones (constants). The weight $W^p_i$ has the same shape as the remaining $p$ higher layers. 

\begin{algorithm}[ht]
	\begin{algorithmic}[1]
		\Require 
		$N$ clients, 
		$\rho$: client joining ratio, 
		$\mathcal{L}$: loss function, 
		$\Theta^{0}$: initial global model, 
		$\alpha$: local learning rate, 
		$\eta$: the learning rate in \module, 
		$s\%$: the percent of local data in \module, 
		$p$: the range of \module, 
		and $\sigma(\cdot)$: clip function.
		\Ensure 
		Reasonable local models $\hat{\Theta}_1, \ldots, \hat{\Theta}_N$ 
        \State Server sends $\Theta^{0}$ to all clients to initialize local models.
        \State Clients initialize $W^p_i, \forall i \in [N]$ to ones.
        \For{iteration $t=1, \ldots, T$}
            \State Server samples a subset $\mathcal{I}^t$ of clients according to $\rho$.
            \State Server sends $\Theta^{t-1}$ to $|\mathcal{I}^t|$ clients. 
		    \For{Client $i \in \mathcal{I}^t$ in parallel}
		        \State Client $i$ samples $s\%$ of local data. \Comment{\textbf{\module}}
	            \If{$t = 2$} \Comment{\textbf{Initial stage}}
		            \While{$W^p_i$ does not converge}
            		    \State Client $i$ trains $W^p_i$ by \Cref{eq:mask_update}.
            		    \State Client $i$ clips $W^p_i$ using $\sigma(\cdot)$.
            		\EndWhile
            	\ElsIf{$t > 2$}
        		    \State Client $i$ trains $W^p_i$ by \Cref{eq:mask_update}.
        		    \State Client $i$ clips $W^p_i$ using $\sigma(\cdot)$.
        		\EndIf
		        \State Client $i$ obtains $\hat{\Theta}^{t}_i$ by \Cref{eq:extract-}.
		        \State Client $i$ obtains $\Theta^{t}_i$ by \Comment{\textbf{Local model training}}
		        \Statex $\Theta^{t}_i \leftarrow \hat{\Theta}^{t}_i - \alpha \nabla_{\hat{\Theta}_i} \mathcal{L}(\hat{\Theta}^{t}_i, D_i; \Theta^{t-1})$.
		        \State Client $i$ sends $\Theta^{t}_i$ to the server. \Comment{\textbf{Uploading}}
		    \EndFor
		    \State Server obtains $\Theta^{t}$ by $\Theta^{t} \leftarrow \sum_{i\in \mathcal{I}^t} \frac{k_i}{\sum_{j\in \mathcal{I}^t} k_j} \Theta^{t}_i$.
		\EndFor
		\\
		\Return $\hat{\Theta}_1, \ldots, \hat{\Theta}_N$
	\end{algorithmic}
	\caption{\method}
	\label{algo}
\end{algorithm}

We initialize the value of each element in $W^p_i$ to one in the beginning and learn $W^p_i$ based on old $W^p_i$ in each iteration. To further reduce computation overhead, we randomly sample s\% of $D_i$ in iteration $t$ and denote it as $D^{s, t}_i$. Client $i$ trains $W^p_i$ through the gradient-based learning method:
\begin{equation}
    W^p_i \leftarrow W^p_i - \eta \nabla_{W^p_i} \mathcal{L}(\hat{\Theta}^{t}_i, D^{s, t}_i; \Theta^{t-1}), \label{eq:mask_update}
\end{equation}
where $\eta$ is the learning rate for weight learning. We freeze other trainable parameters in \module, including the entire global model and entire local model. 
After local initialization, client $i$ performs local model training as in FedAvg. 

\begin{figure}[ht]
	\centering
	\includegraphics[width=\linewidth]{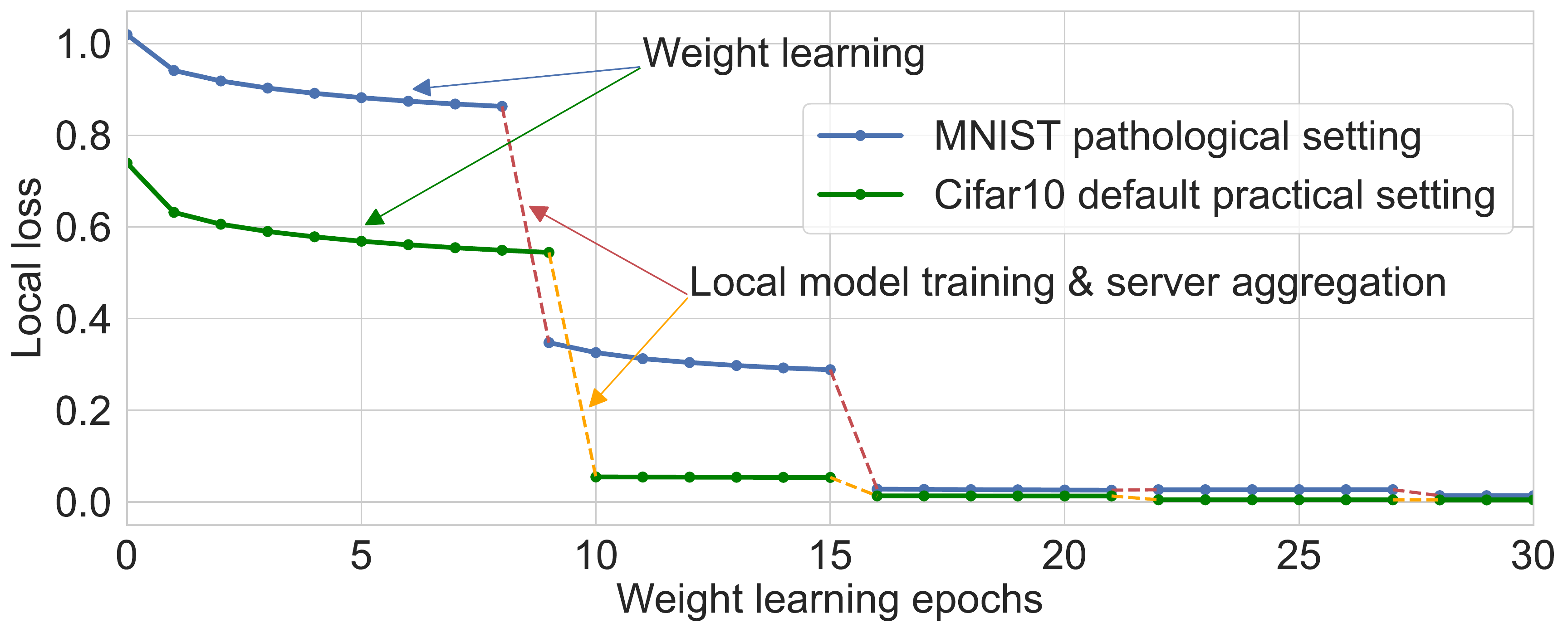}
	\caption{The local loss on client \#8 regarding weight learning epochs in \module on MNIST and Cifar10. Here, we train the weights for at least six epochs in each iteration. The details of the settings are described in Section \ref{sec:exp}. }
	\label{fig:wl}
\end{figure}

We can significantly reduce the number of the trainable parameters in \module by choosing a small $p$ with negligible performance decline. We show the details in Section \ref{sec:hyper}. 
Besides, we observe that once we train $W^p_i$ to converge in the second iteration (initial stage), it hardly changes in the subsequent iterations, as shown in \Cref{fig:wl}. In other words, $W^p_i$ can be reused. Similar to APPLE and PartialFed, we train only one epoch for $W^p_i$ when $t>2$ to adapt to the changing model parameters. Note that \module is meaningless and deactivated in the first iteration since $\Theta^{0}=\Theta^{0}_i, \forall i\in [N]$. 
\Cref{algo} presents the overall FL process in \method. 

\subsection{Analysis of \module}
\label{sec:theoretical}
We omit $\sigma(\cdot)$ and set $p=|\Theta_i|$ here for simplicity without affecting the analysis. According to \Cref{eq:extract-} and \Cref{eq:mask_update}, we have $\nabla_{W_i} \mathcal{L}^{t}_i = \eta (\Theta^{t-1} - \Theta^{t-1}_i) \odot \nabla_{\hat{\Theta}_i} \mathcal{L}^{t}_i$, where $\mathcal{L}^{t}_i$ represents $\mathcal{L}(\hat{\Theta}^{t}_i, D^{s, t}_i; \Theta^{t-1})$. Based on \Cref{eq:extract-}, we can view updating $W_i$ as updating $\hat{\Theta}^{t}_i$ in \module:
\begin{equation}
    \hat{\Theta}^{t}_i \leftarrow  \hat{\Theta}^{t}_i - \eta (\Theta^{t-1} - \Theta^{t-1}_i) \odot (\Theta^{t-1} - \Theta^{t-1}_i) \odot \nabla_{\hat{\Theta}_i} \mathcal{L}^{t}_i. \label{eq:ele}
\end{equation}
The gradient $\nabla_{\hat{\Theta}_i} \mathcal{L}^{t}_i$ is element-wisely scaled in iteration $t$. In contrast to local model training (or fine-tuning) that only focuses on the local data, the whole process of updating in \Cref{eq:ele} can be aware of the generic information in the global model. Across iterations, the dynamic term $(\Theta^{t-1} - \Theta^{t-1}_i)$ brings dynamic information for \module to adapt to the complex scenarios, although it is a constant in iteration $t$.

\section{Experiments}
\label{sec:exp}

\begin{table*}[ht]
  \centering
  \resizebox{\linewidth}{!}{
    \begin{tabular}{l|*{7}{c}|*{6}{c}}
    \toprule
    \multirow{2}{*}{Items} & \multicolumn{7}{c|}{$p=1$} & \multicolumn{6}{c}{$s=80$} \\ 
    \cmidrule{2-14}
    & $s=5$ & $s=10$ & $s=20$ & $s=40$ & $s=60$ & $s=80$ & $s=100$ & $p=6$ & $p=5$ & $p=4$ & $p=3$ & $p=2$ & $p=1$\\ 
    \midrule
    Acc. & 39.53 & 40.62 & 40.02 & 40.23 & 41.11 & 41.94 & \textbf{42.11} & 41.71 & 41.54 & 41.62 & 41.86 & \textbf{42.47} & 41.94 \\
    \midrule
    Param. & \multicolumn{7}{c|}{0.005} & 11.182 & 11.172 & 11.024 & 10.499 & 8.399 & 0.005 \\
    \bottomrule
    \end{tabular}}
  \caption{The test accuracy (\%) and the number of trainable parameters (in millions) for \module on TINY*.}
  \label{tab:s}
\end{table*}

\begin{table*}[ht]
  \centering
  \resizebox{\linewidth}{!}{
    \begin{tabular}{l|*{3}{c}|*{5}{c}}
    \toprule
    Settings & \multicolumn{3}{c|}{Pathological heterogeneous setting} & \multicolumn{5}{c}{Practical heterogeneous setting} \\
    \midrule
    Methods & MNIST & Cifar10 & Cifar100 & Cifar10 & Cifar100 & TINY & TINY* & AG News\\
    \midrule
    FedAvg & 97.93$\pm$0.05 & 55.09$\pm$0.83 & 25.98$\pm$0.13 & 59.16$\pm$0.47 & 31.89$\pm$0.47 & 19.46$\pm$0.20 & 19.45$\pm$0.13 & 79.57$\pm$0.17\\
    FedProx & 98.01$\pm$0.09 & 55.06$\pm$0.75 & 25.94$\pm$0.16 & 59.21$\pm$0.40 & 31.99$\pm$0.41 & 19.37$\pm$0.22 & 19.27$\pm$0.23 & 79.35$\pm$0.23\\
    \midrule
    FedAvg-C & 99.79$\pm$0.00 & 92.13$\pm$0.03 & 66.17$\pm$0.03 & 90.34$\pm$0.01 & 51.80$\pm$0.02 & 30.67$\pm$0.08 & 36.94$\pm$0.10 & 95.89$\pm$0.25\\
    FedProx-C & 99.80$\pm$0.04 & 92.12$\pm$0.03 & 66.07$\pm$0.08 & 90.33$\pm$0.01 & 51.84$\pm$0.07 & 30.77$\pm$0.13 & 38.78$\pm$0.52 & 96.10$\pm$0.22\\
    \midrule
    Per-FedAvg & 99.63$\pm$0.02 & 89.63$\pm$0.23 & 56.80$\pm$0.26 & 87.74$\pm$0.19 & 44.28$\pm$0.33 & 25.07$\pm$0.07 & 21.81$\pm$0.54 & 93.27$\pm$0.25\\
    FedRep & 99.77$\pm$0.03 & 91.93$\pm$0.14 & 67.56$\pm$0.31 & 90.40$\pm$0.24 & 52.39$\pm$0.35 & 37.27$\pm$0.20 & 39.95$\pm$0.61 & 96.28$\pm$0.14\\
    pFedMe & 99.75$\pm$0.02 & 90.11$\pm$0.10 & 58.20$\pm$0.14 & 88.09$\pm$0.32 & 47.34$\pm$0.46 & 26.93$\pm$0.19 & 33.44$\pm$0.33 & 91.41$\pm$0.22\\
    Ditto & 99.81$\pm$0.00 & 92.39$\pm$0.06 & 67.23$\pm$0.07 & 90.59$\pm$0.01 & 52.87$\pm$0.64 & 32.15$\pm$0.04 & 35.92$\pm$0.43 & 95.45$\pm$0.17\\
    FedAMP & 99.76$\pm$0.02 & 90.79$\pm$0.16 & 64.34$\pm$0.37 & 88.70$\pm$0.18 & 47.69$\pm$0.49 & 27.99$\pm$0.11 & 29.11$\pm$0.15 & 94.18$\pm$0.09\\
    FedPHP & 99.73$\pm$0.00 & 90.01$\pm$0.00 & 63.09$\pm$0.04 & 88.92$\pm$0.02 & 50.52$\pm$0.16 & 35.69$\pm$3.26 & 29.90$\pm$0.51 & 94.38$\pm$0.12\\
    FedFomo & 99.83$\pm$0.00 & 91.85$\pm$0.02 & 62.49$\pm$0.22 & 88.06$\pm$0.02 & 45.39$\pm$0.45 & 26.33$\pm$0.22 & 26.84$\pm$0.11 & 95.84$\pm$0.15\\
    APPLE & 99.75$\pm$0.01 & 90.97$\pm$0.05 & 65.80$\pm$0.08 & 89.37$\pm$0.11 & 53.22$\pm$0.20 & 35.04$\pm$0.47 & 39.93$\pm$0.52 & 95.63$\pm$0.21\\
    PartialFed & 99.86$\pm$0.01 & 89.60$\pm$0.13 & 61.39$\pm$0.12 & 87.38$\pm$0.08 & 48.81$\pm$0.20 & 35.26$\pm$0.18 & 37.50$\pm$0.16 & 85.20$\pm$0.16\\
    \midrule
    \method & \textbf{99.88$\pm$0.01} & \textbf{92.44$\pm$0.02} & \textbf{67.83$\pm$0.06} & \textbf{90.67$\pm$0.03} & \textbf{55.92$\pm$0.03} & \textbf{40.54$\pm$0.02} & \textbf{41.94$\pm$0.05} & \textbf{96.52$\pm$0.08}\\
    \bottomrule
    \end{tabular}}
  \caption{The test accuracy (\%) in the pathological heterogeneous setting and practical heterogeneous setting.}
    \label{tab:pathological}
\end{table*}

\subsection{Experimental Setup}

In this section, we firstly compare \method with eleven SOTA FL baselines including FedAvg~\cite{mcmahan2017communication}, FedProx~\cite{MLSYS2020_38af8613}, Per-FedAvg~\cite{NEURIPS2020_24389bfe}, FedRep~\cite{collins2021exploiting}, pFedMe~\cite{t2020personalized}, Ditto~\cite{li2021ditto}, FedAMP~\cite{huang2021personalized}, FedPHP~\cite{li2021fedphp}, FedFomo~\cite{zhang2020personalized}, APPLE~\cite{luo2021adapt}, and PartialFed~\cite{sun2021partialfed}. To show the superiority of weight learning in \method over additional local training steps, we also compare \method with FedAvg-C and FedProx-C, which locally fine-tune the global model for local initialization at each iteration. Then we apply \module to SOTA FL methods and show that \module can improve them. We conduct extensive experiments in computer vision (CV) and natural language processing (NLP) domains. 

For the CV domain, we study the image classification tasks with four widely used datasets including MNIST~\cite{lecun1998gradient}, Cifar10/100~\cite{krizhevsky2009learning} and Tiny-ImageNet~\cite{chrabaszcz2017downsampled} (100K images with 200 classes) using the 4-layer CNN~\cite{mcmahan2017communication}. For the NLP domain, we study the text classification tasks with AG News~\cite{zhang2015character} and fastText~\cite{joulinetal2017bag}. To evaluate the effectiveness of \method on a larger model, we also use ResNet-18~\cite{he2016deep} on Tiny-ImageNet. We set the local learning rate to 0.005 for the 4-layer CNN (0.1 on MNIST following FedAvg) and 0.1 for both fastText and ResNet-18. We set the batch size to 10 and the number of local model training epochs to 1, following FedAvg. We run all the tasks for 2000 iterations to make all the methods converge empirically. Following pFedMe, we have 20 clients and set $\rho=1$ by default. 

We simulate the heterogeneous settings with two widely used scenarios. The first one is the pathological heterogeneous setting~\cite{mcmahan2017communication, pmlrv139shamsian21a}, where we sample 2/2/10 classes for MNIST/Cifar10/Cifar100 from a total of 10/10/100 classes for each client, with disjoint data samples. The second scenario is the practical heterogeneous setting~\cite{NEURIPS2020_18df51b9, li2021model}, which is controlled by the Dirichlet distribution, denoted as $Dir(\beta)$. The smaller the $\beta$ is, the more heterogeneous the setting is. We set $\beta=0.1$ for the default heterogeneous setting~\cite{NEURIPS2020_18df51b9, NEURIPS2020_564127c0}. 

We use the same evaluation metrics as pFedMe, which reports the test accuracy of the best single global model for the traditional FL and the average test accuracy of the best local models for pFL. To simulate the practical pFL setting, we evaluate the learned model on the client side. 25\% of the local data forms the test dataset, and the remaining 75\% data is used for training. We run all the tasks five times and report the mean and standard deviation. 

We implement \method using PyTorch-1.8 and run all experiments on a server with two Intel Xeon Gold 6140 CPUs (36 cores), 128G memory, and eight NVIDIA 2080 Ti GPUs, running CentOS 7.8.

\subsection{Effect of Hyperparameters}
\label{sec:hyper}

\subsubsection{Effect of $s$.} From \Cref{tab:s}, high accuracy corresponds to large $s$, where ``TINY*'' represents using ResNet-18 on Tiny-ImageNet in the default heterogeneous setting. We can balance the performance and the computational cost by choosing a reasonable value for $s$. \method can also achieve excellent performance with only 5\% local data for \module. Since the improvement from $s=80$ to $s=100$ is negligible, we set $s=80$ for \method.

\subsubsection{Effect of $p$.} By decreasing the hyperparameter $p$, we can shrink the range of \module with negligible accuracy decrease, as shown in \Cref{tab:s}. When $p$ decreases from 6 to 1, the number of trainable parameters in \module also decreases, especially from $p=2$ to $p=1$, as the last block in ResNet-18 contains most of the parameters~\cite{he2016deep}. Although \method performs the best when $p=2$ here, we set $p=1$ for ResNet-18 to reduce computation overhead. Similarly, we also set $p=1$ for the 4-layer CNN and fastText. This also shows that the lower layers of the global model mostly contain generic information which is desired by the client. 

\begin{table}[ht]
  \centering
  \resizebox{\linewidth}{!}{
    \begin{tabular}{l|rr|c}
    \toprule
    & \multicolumn{2}{c|}{Computation} & Communication\\
    \midrule
    & Total time & Time/iter. & Param./iter. \\
    \midrule
    FedAvg & 365 min & 1.59 min & $2 * \Sigma$ \\
    FedProx & 325 min & 1.99 min & $2 * \Sigma$\\
    \midrule
    FedAvg-C & 607 min & 24.28 min & $2 * \Sigma$\\
    FedProx-C & 711 min & 28.44 min & $2 * \Sigma$\\
    \midrule
    Per-FedAvg & 121 min & 3.56 min & $2 * \Sigma$\\
    FedRep & 471 min & 4.09 min & $2 * \alpha_f * \Sigma$\\
    pFedMe & 1157 min & 10.24 min & $2 * \Sigma$\\
    Ditto & 318 min & 11.78 min & $2 * \Sigma$\\
    FedAMP & 92 min & 1.53 min & $2 * \Sigma$\\
    FedPHP & 264 min & 4.06 min & $2 * \Sigma$\\
    FedFomo & 193 min & 2.72 min & $(1 + M) * \Sigma$\\
    APPLE & 132 min& 2.93 min & $(1 + M) * \Sigma$\\
    PartialFed & 693 min & 2.13 min & $2 * \Sigma$\\
    \midrule
    \method & 7+116 min & 1.93 min & $2 * \Sigma$\\
    \bottomrule
    \end{tabular}}
  \caption{The computation overhead on TINY* and the communication overhead (transmitted parameters per iteration). $\Sigma$ is the parameter amount in the backbone. $\alpha_f$ ($\alpha_f < 1$) is the ratio of the parameters of the feature extractor in the backbone. $M$ ($M \ge 1$) is the number of the received client models on each client in FedFomo and APPLE. }
    \label{tab:com}
\end{table}

\subsection{Performance Comparison and Analysis}
\label{sec:performance}

\subsubsection{Pathological heterogeneous setting.} 
Clients are separated into groups, which benefits the methods that measure the similarity among clients, such as FedAMP. 
Even so, \Cref{tab:pathological} shows that \method outperforms all the baselines. Due to the poor generalization ability of the global model, FedAvg and FedProx perform poorly in this setting. 

\noindent\textbf{pFL methods in Category (1).} Compared to the traditional FL methods, the personalized methods perform better. The accuracy for Per-FedAvg is the lowest among these methods since it only finds an initial shared model corresponding to the learning trends of all the clients, which may not capture the needs of an individual client. Fine-tuning the global model in FedAvg-C/FedProx-C generates client-specific local models, which improves the accuracy for FedAvg/FedProx. However, fine-tuning only focuses on the local data and cannot be aware of the generic information during local training like \method. Although FedRep also fine-tunes the head at each iteration, it freezes the downloaded representation part when fine-tuning and keeps most of the generic information in the global model, thus performing excellently. However, the generic information of the head part is lost without sharing the head among clients. 

\noindent\textbf{pFL methods in Category (2).} Although both pFedMe and Ditto use the proximal term to learn their additional personalized models, pFedMe learns the desired information from the local model while Ditto learns it from the global model. Thus, Ditto learns more generic information locally, and it performs better. However, learning the personalized model with the proximal term is an implicit way to extract the desired information. 

\noindent\textbf{pFL methods in Category (3).} Aggregating the models using the rule-based method is aimless in capturing the desired information in the global model, so FedPHP performs worse than FedRep and Ditto. The model-level personalized aggregation in FedAMP, FedFomo, and APPLE, as well as the layer-level and binary selection in PartialFed, are imprecise, which may introduce undesired information in the global model to the local model. Furthermore, downloading multiple models on each client in each iteration has additional communication costs for FedFomo and APPLE. 

By adaptively learning the aggregating weights, \method can explicitly capture the desired information precisely in the global model to facilitate the local model training. 

\subsubsection{Practical heterogeneous setting.}
We add two additional tasks using Tiny-ImageNet and one text classification task on AG News in the default practical heterogeneous setting. The results in the default heterogeneous setting with $Dir(0.1)$ are shown in \Cref{tab:pathological}, where ``TINY'' represents using the 4-layer CNN on Tiny-ImageNet. 
\method is still superior to other baselines, which achieves an accuracy of 3.27\% higher than FedRep on TINY. 

In the practical heterogeneous setting, due to the complex data distribution of each client, it is hard to measure the similarity among clients. Thus, FedAMP can not precisely assign importance to the local models through the attention-inducing function to generate aggregated models with personalized aggregation.
After downloading the global model/representation, Ditto and FedRep can capture the generic information from it instead of measuring the similarity among local models. In this way, they achieve excellent performance in most of the tasks. The trainable weights are more informative than the approximated one, so APPLE performs better than FedFomo. 
Although FedPHP performs well on TINY, the standard deviation is relatively high. 
Since \method can adapt to the changing circumstances through \module, it still outperforms all the baselines in the practical setting. If we reuse the aggregated weights learned in the initial stage without adaptation, the accuracy drops to 33.81\% on TINY. Due to the fine-grained feature of \module, it still outperforms Per-FedAvg, pFedMe, Ditto, FedAMP, and FedFomo. 

Compared to the 4-layer CNN, ResNet-18 is a larger backbone. With ResNet-18, most methods achieve a higher accuracy, including \method. However, it is harder for FedAMP to measure the model similarity, and the heuristic local aggregation in FedPHP performs worse when using ResNet-18. As we set $p=1$, the number of trainable parameters in \module is much less than that of ResNet-18, but \method can still achieve superior performance. 

\begin{table*}[ht]
  \centering
  \resizebox{\linewidth}{!}{
    \begin{tabular}{l|*{3}{c}|*{2}{c}|*{2}{c}|*{2}{c}}
    \toprule
    & \multicolumn{3}{c|}{Heterogeneity} & \multicolumn{2}{c|}{Scalability} & \multicolumn{4}{c}{Applicability of \module} \\
    \midrule
    Datasets & \multicolumn{2}{c}{Tiny-ImageNet} & AG News & \multicolumn{2}{c|}{Cifar100} & \multicolumn{2}{c|}{Tiny-ImageNet} & \multicolumn{2}{c}{Cifar100} \\
    \midrule
    Methods & $Dir(0.01)$ & $Dir(0.5)$ & $Dir(1)$ & 50 clients & 100 clients & Acc. & Imps. & Acc. & Imps.\\
    \midrule
    FedAvg & 15.70$\pm$0.46 & 21.14$\pm$0.47 & 87.12$\pm$0.19 & 31.90$\pm$0.27 & 31.95$\pm$0.37 & 40.54$\pm$0.17 & 21.08 & 55.92$\pm$0.15 & 24.03\\
    FedProx & 15.66$\pm$0.36 & 21.22$\pm$0.47 & 87.21$\pm$0.13 & 31.94$\pm$0.30 & 31.97$\pm$0.24 & 40.53$\pm$0.26 & 21.16 & 56.18$\pm$0.65 & 24.19 \\
    \midrule
    FedAvg-C & 49.88$\pm$0.11 & 16.21$\pm$0.05 & 91.38$\pm$0.21 & 49.82$\pm$0.11 & 47.90$\pm$0.12 & --- & --- & --- & ---\\
    FedProx-C & 49.84$\pm$0.02 & 16.36$\pm$0.19 & 92.03$\pm$0.19 & 49.79$\pm$0.14 & 48.02$\pm$0.02 & --- & --- & --- & ---\\
    \midrule
    Per-FedAvg & 39.39$\pm$0.30 & 16.36$\pm$0.13 & 87.08$\pm$0.26 & 44.31$\pm$0.20 & 36.07$\pm$0.24 & 30.90$\pm$0.28 & 5.83 & 48.68$\pm$0.36 & 4.40\\
    FedRep & 55.43$\pm$0.15 & 16.74$\pm$0.09 & 92.25$\pm$0.20 & 47.41$\pm$0.18 & 44.61$\pm$0.20 & 37.89$\pm$0.31 & 0.62 & 53.02$\pm$0.11 & 0.63\\
    pFedMe & 41.45$\pm$0.14 & 17.48$\pm$0.61 & 87.08$\pm$0.18 & 48.36$\pm$0.64 & 46.45$\pm$0.18  & 27.30$\pm$0.24 & 0.37 & 47.91$\pm$0.21 & 0.57\\
    Ditto & 50.62$\pm$0.02 & 18.98$\pm$0.05 & 91.89$\pm$0.17 & 54.22$\pm$0.04 & 52.89$\pm$0.22 & 40.75$\pm$0.06 & 8.60 & 56.33$\pm$0.07 & 3.46\\
    FedAMP & 48.42$\pm$0.06 & 12.48$\pm$0.21 & 83.35$\pm$0.05 & 44.39$\pm$0.35 & 40.43$\pm$0.17  & 28.18$\pm$0.20 & 0.19 & 48.03$\pm$0.23 & 0.34\\
    FedPHP & 48.63$\pm$0.02 & 21.09$\pm$0.07 & 90.52$\pm$0.19 & 52.44$\pm$0.16 & 49.70$\pm$0.31 & 40.16$\pm$0.24 & 4.47 & 54.28$\pm$0.21 & 3.76\\
    FedFomo & 46.36$\pm$0.54 & 11.59$\pm$0.11 & 91.20$\pm$0.18 & 42.56$\pm$0.33 & 38.91$\pm$0.08 & --- & --- & --- & ---\\
    APPLE & 48.04$\pm$0.10 & 24.28$\pm$0.21 & 84.10$\pm$0.18 & 55.06$\pm$0.20 & 52.81$\pm$0.29 & --- & --- & --- & ---\\
    PartialFed & 49.38$\pm$0.02 & 24.20$\pm$0.10 & 91.01$\pm$0.28 & 48.95$\pm$0.07 & 39.31$\pm$0.01 & 35.40$\pm$0.02 & 0.14 & 48.99$\pm$0.05 & 0.18\\
    \midrule
    \method & \textbf{55.75$\pm$0.02} & \textbf{27.85$\pm$0.06} & \textbf{92.45$\pm$0.10} & \textbf{55.61$\pm$0.02} & \textbf{54.68$\pm$0.57} &--- & --- & --- & ---\\
    \bottomrule
    \end{tabular}}
  \caption{The test accuracy (\%) (and improvement (\%)) on Tiny-ImageNet, AG News, and Cifar100.}
    \label{tab:beta}
\end{table*}

\subsubsection{Computation Overhead.}
We record the total time cost for each method until convergence, as shown in \Cref{tab:com}. Except for the 7 min cost in the initial stage, \method costs 1.93 min (similar to FedAvg) in each iteration. In other words, \module only costs an additional 0.34 min for the great accuracy improvement. 
However, FedAvg-C, FedProx-C, and FedRep cost relatively more time than most of the methods because of the fine-tuning for the entire local model (or only the head). Due to the additional training steps for the personalized model, Ditto and pFedMe have the top computation overhead per iteration among SOTA methods. FedFomo and APPLE feed data forward in the downloaded client-side model to approximate aggregate weights and update directed relation (DR) vectors, respectively, taking additional time.

\subsubsection{Communication Overhead.}

We show the communication overhead for one client in one iteration in \Cref{tab:com}. The communication overhead for most of the methods is the same as FedAvg, which uploads and downloads only one model. 
Since FedRep only transmits the representation and keeps the head locally, it has less communication overhead. 
FedFomo and APPLE require the highest communication overhead, as they download $M$ client models in each iteration~\cite{zhang2020personalized, luo2021adapt}. To achieve the results in \Cref{tab:pathological}, we set $M=20$ for them. 

\subsubsection{Heterogeneity.}
To study the effectiveness of \method in the settings with different degrees of heterogeneity, we vary the $\beta$ in $Dir(\beta)$ on Tiny-ImageNet and AG News. The smaller the $\beta$ is, the more heterogeneous the setting is. In \Cref{tab:beta}, most of the pFL methods have better performance in the more heterogeneous settings. However, except for FedPHP, APPLE, PartialFed, and \method, these methods excessively focus on personalization but underestimate the significance of the generic information. When the data heterogeneity becomes moderate with $Dir(0.5)$ for Tiny-ImageNet, they perform worse than FedAvg.

\subsubsection{Scalability.}

To show the scalability of \method, we conduct two experiments with 50 and 100 clients in the default heterogeneous setting. In \Cref{tab:beta}, most of the pFL methods degenerate greatly as the client amount increases to 100, while \method drops less than 1\% in accuracy. Since the data amount on Cifar100 is constant, the data amount on the client decreases with more clients. This aggravates the lack of local data, so precisely capturing the desired information in the global model becomes more critical.

\subsection{Applicability of \module}

As the \module module only modifies the local initialization in FL, it can be applied to most existing FL methods. We apply \module to the SOTA FL methods except for FedFomo and APPLE (as they download multiple client models) without modifying other learning processes to evaluate the effectiveness of \module. We report the accuracy and improvements on Tiny-ImageNet and Cifar100 using the 4-layer CNN in default heterogeneous setting with $s=80$ and $p=1$. 

In \Cref{tab:beta}, the accuracy improvement for FedAvg and FedProx is apparent, and the improvement for Per-FedAvg, Ditto, and FedPHP is also remarkable. This indicates the applicability of \module to the traditional FL and pFL methods. 

However, the improvement to other pFL methods is relatively small. 
In FedAMP, only the information in the important local models is emphasized through the attention-inducing function with the generic information in unimportant models ignored, so \module can hardly find the ignored information back without modifying other learning processes.
According to \Cref{tab:s}, the representation part mostly contains generic information, which is desired by the client. It leaves little room for \module to take effect, but \module still improves FedRep by more than 0.60\% in accuracy. pFedMe learns a personalized model to approximate the local model at each local training batch, so it benefits little (0.57\% on Cifar100) from \module. In PartialFed, the local aggregation happens in each training batch, so the initialized local model generated by \module is later re-aggregated by its strategy, thus eliminating the effect of \module. 

\begin{figure}[t]
	\centering
	\includegraphics[width=\linewidth]{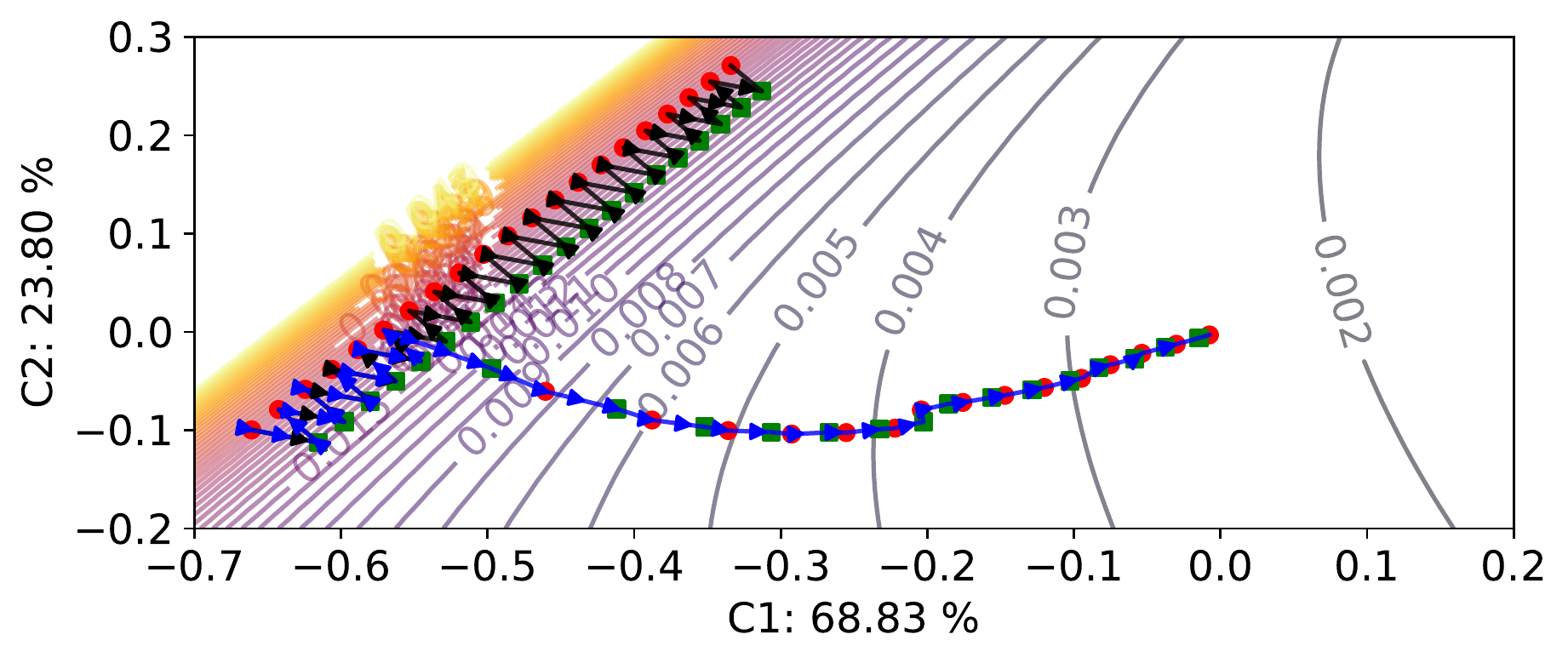}
	\caption{2D visualization of local learning trajectory (from iteration 140 to 200) and the local loss surface on MNIST in the pathological heterogeneous setting. 
	The red circles and green cubes represent the local model at the beginning and end of each iteration, respectively. 
	The black and blue trajectories with the arrows represent FedAvg and \method, respectively. 
	The local models are projected to the 2D plane using PCA. 
    C1 and C2 are the two principal components generated by the PCA.
	}
    \label{fig:traj}
\end{figure}

\subsection{Update Direction Correction}
\label{sec:direction}

Compared to overwriting the local model with the downloaded global model, \module can correct the update direction for the local model with the desired information in the global model. We visualize the learning trajectory of the local model on client \#4 using the visualization method~\cite{li2018visualizing}.
We deactivate the \module for \method in the first 155 iterations and activate it in the subsequent iterations.

Without capturing the desired information in the global model, the update direction of local model is misled by the global model in FedAvg, as shown by the black trajectory in \Cref{fig:traj}. Once the \module is activated, the update direction of local model is corrected to the local loss reducing direction, as shown by the blue trajectory in \Cref{fig:traj}.

\section{Conclusion}

In this paper, we propose an adaptive and fine-grained method \method in pFL to facilitate the local model training with the received global model. The extensive experiments demonstrate the effectiveness of \method. Our method outperforms eleven SOTA methods. Also, the \module module in \method can improve other FL methods in accuracy. 

\section*{Acknowledgements}

This work was supported in part by National NSF of China (NO. 61872234, 61732010), Shanghai Key Laboratory of Scalable Computing and Systems, Innovative Research Foundation of Ship General Performance (NO.25622114), the Key Laboratory of PK System Technologies Research of Hainan, Intel Corporation (UFunding 12679), and the Louisiana BoR LAMDA.

\bibliography{JianqingZhang}

\appendix
\section{Additional Details of weight learning in \module}

Here, we omit the superscript $t$ for all the notations. Given $\hat{\Theta}_i := \Theta_i + (\Theta - \Theta_i) \odot [\textbf{1}^{|\Theta_i|-p}; W^{p}_i]$, we update $W^{p}_i$ by \Cref{eq:mask_update_detail} shown as below with other learnable weights frozen, including the entire global model and entire local model. 

\begin{equation}
\begin{aligned}
    W^p_i &\leftarrow W^p_i - \eta \nabla_{W^p_i} \mathcal{L}(\hat{\Theta}_i, D^{s}_i; \Theta) \\
    &= W^p_i - \eta \frac{\partial \mathcal{L}_i}{\partial W^p_i} \\
    &= W^p_i - \eta \frac{\partial \mathcal{L}_i}{\partial \hat{\Theta}^p_i} \odot \frac{\partial \hat{\Theta}^p_i}{\partial W^p_i}\\
    &= W^p_i - \eta \frac{\partial \mathcal{L}_i}{\partial \hat{\Theta}^p_i} \odot (\Theta - \Theta_i)^p,
\end{aligned}\label{eq:mask_update_detail}
\end{equation}
where $\mathcal{L}_i$ represents $\mathcal{L}(\hat{\Theta}_i, D^{s}_i; \Theta)$, $\hat{\Theta}^p_i$ and $(\Theta - \Theta_i)^p$ represent the higher layers of $\hat{\Theta}_i$ and $(\Theta - \Theta_i)$, respectively. The term $\frac{\partial \mathcal{L}_i}{\partial \hat{\Theta}^p_i}$ can be easily obtained with the backpropagation. Only the gradients of the model parameters in the higher layers are calculated in \module. 

\section{Convergence of \method}

\begin{figure*}[ht]
	\centering
	\includegraphics[width=0.9\linewidth]{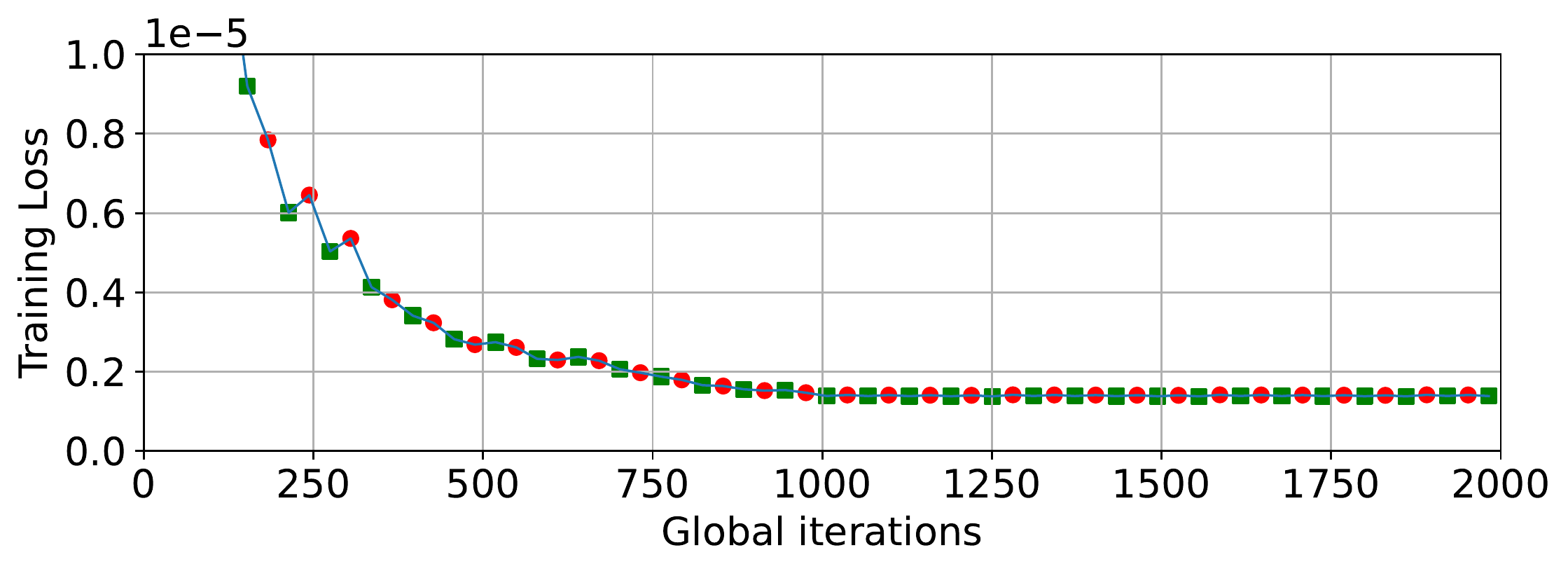}
	\caption{The training loss of the global objective (\Cref{eq:gobj}) in \method on MNIST in the pathological setting. We alternatively record the averaged losses of the trained local models (green square dots) and the averaged losses of the initialized local model before local training (red circle dots) in each iteration. For clarity, we show one dot per 60 rounds. Note the magnitude of y-axis. }
	\label{fig:con}
\end{figure*} 

Recall that our global objective is

\begin{equation}
    \{\hat{\Theta}_1, \ldots, \hat{\Theta}_N\} = \argmin \ \mathcal{G}(\mathcal{L}_1, \ldots, \mathcal{L}_N), \label{eq:gobj}
\end{equation}
where $\mathcal{L}_i = \mathcal{L}(\hat{\Theta}_i, D_i; \Theta), \forall i\in [N]$ and $\mathcal{L}(\cdot)$ is the loss function. $\Theta$ is the global model, which brings external information to client $i$. Typically $\mathcal{G}(\cdot)$ is set to $\sum^{N}_{i=1} k_i \mathcal{L}_i$, where $k_i = |D_i| / \sum^{N}_{j=1} |D_j|$ and $|D_i|$ is the amount of local training samples on client $i$. 

As shown in \Cref{fig:con}, the training loss value of the global objective keeps decreasing when considering either the trained local models (green square dots) or the initialized local models before local training (red circle dots). When the number of global iterations becomes larger than 1000, the loss of the green dots and the red dots become almost the same, representing the convergence of \method. 

\section{The Range of \module}

For different backbones, the ranges of \module with different $p$ are shown in \Cref{tab:range-resnet}. We use the notations from the corresponding papers. As $p$ increases, the range increases, so the \module module can cover more layers. 

\begin{table*}[ht]
	\caption{The Range of \module with different $p$. We use the notations in~\cite{he2016deep, mcmahan2017communication, joulinetal2017bag}.}
	\centering
	\resizebox{\textwidth}{!}{
	\begin{tabular}{l*{3}{|r}}
		\toprule
		$p$ & ResNet-18 & Four-layer CNN & fastText \\
		\midrule
		$p=1$ & fc & output & output\\
		$p=2$ & conv5\_x+fc & fc+output & hidden+output\\
		$p=3$ & conv4\_x+conv5\_x+fc & conv\_2+fc+output & embedding+hidden+output\\
		$p=4$ & conv3\_x+conv4\_x+conv5\_x+fc & conv\_1+conv\_2+fc+output & \multicolumn{1}{|c}{/}\\
		$p=5$ & conv2\_x+conv3\_x+conv4\_x+conv5\_x+fc & \multicolumn{1}{|c}{/} & \multicolumn{1}{|c}{/}\\
		$p=6$ & conv1+conv2\_x+conv3\_x+conv4\_x+conv5\_x+fc & \multicolumn{1}{|c}{/} & \multicolumn{1}{|c}{/}\\
		\bottomrule
	\end{tabular}}\label{tab:range-resnet}
\end{table*}

\section{Effect of $p$ on CNN and fastText}

With different $p$, the accuracy of \method and the number of learnable weights in \module also varies. In \Cref{tab:p-cnn}, as $p$ increases, the number of learnable weights also increases. However, the change in the accuracy is negligible. The accuracy reaches the best for the four-layer CNN and fastText when $p=1$. 

\begin{table*}[ht]
  \centering
  \caption{The test accuracy (\%) and the number of learnable weights in \module module on Tiny-ImageNet in the default heterogeneous setting. $s$ is set to 80\%.}\label{tab:p-cnn}
  \resizebox{!}{!}{
    \begin{tabular}{l|*{4}{c}|*{3}{c}}
    \toprule
    \multirow{2}{*}{Items} & \multicolumn{4}{c|}{Four-layer CNN} & \multicolumn{3}{c}{fastText} \\
    \cmidrule{2-8}
     & $p=4$ & $p=3$ & $p=2$ & $p=1$ & $p=3$ & $p=2$ & $p=1$ \\
    \midrule
    Acc. & 39.75 & 39.89 & 39.94 & 40.54 & 96.45 & 96.40 & 96.52 \\
    Param. & 582026 & 581194 & 529930 & 5130 & 3157508 & 1188 & 132 \\
    \bottomrule
    \end{tabular}}
\end{table*}

\begin{table*}[ht]
  \centering
  \caption{The test accuracy (\%) using the four-layer CNN on MNIST and Cifar100 in the practical setting with $\beta=0.1$}
  \resizebox{!}{!}{
    \begin{tabular}{l|c|cc}
    \toprule
    Datasets & MNIST & Cifar100\\
    \midrule
    Client Amount & 20 & 100 \\
    \midrule
     & $\rho=1$ & $\rho=0.5^*$ \\
    \midrule
    FedAvg & 98.81$\pm$0.01 & 39.51$\pm$1.23 \\
    FedProx & 98.82$\pm$0.01 & 33.87$\pm$2.39 \\
    \midrule
    FedAvg-C & 99.65$\pm$0.00 & 47.94$\pm$0.26 \\
    FedProx-C & 99.64$\pm$0.00 & 48.11$\pm$0.17 \\
    \midrule
    Per-FedAvg & 98.90$\pm$0.05 & 47.96$\pm$0.83 \\
    FedRep & 99.48$\pm$0.02 & 41.48$\pm$0.05 \\
    pFedMe & 99.52$\pm$0.02 & 43.27$\pm$0.46 \\
    Ditto & 99.64$\pm$0.00 & 48.94$\pm$0.04 \\
    FedAMP & 99.47$\pm$0.02 & --- \\
    FedPHP & 99.58$\pm$0.00 & 49.99$\pm$0.73 \\
    FedFomo & 99.33$\pm$0.04 & 37.70$\pm$0.10 \\
    APPLE & 99.66$\pm$0.02 & --- \\
    PartialFed & 99.67$\pm$0.01 & 36.49$\pm$0.07 \\
    \midrule
    \method & \textbf{99.71$\pm$0.00} & \textbf{54.81$\pm$0.03} \\
    \bottomrule
    \end{tabular}}
    \label{tab:mnist}
\end{table*}

\section{Additional Results on MNIST}

In addition to the results shown in Table 2 in the main body, we present the results on MNIST in the practical heterogeneous setting here, as shown in \Cref{tab:mnist}. According to \Cref{tab:mnist}, \method still outperforms all the baselines on MNIST. 

\section{Additional Results on Cifar100 with $\rho=0.5$}

Following pFedMe, we have shown the superiority of the \method on extensive experiments in the main body with $\rho=1.0$. Here, we conduct additional experiments on Cifar100 (100 clients) with $\rho=0.5$, \ie, only half of the clients randomly join FL in each iteration. Besides, we only show the averaged results collected from the joining clients and denote it as $\rho=0.5^*$ in \Cref{tab:mnist}. FedAMP and APPLE are proposed for the cross-silo FL setting, and they require all clients to join each iteration. Thus, we do not compare them with other methods when $\rho=0.5$. Attributed to the adaptive module \module, \method maintains its superiority. 

\section{Hyperparameter Settings}

We tune all the hyperparameters in the default practical setting by grid search (the search range is included in [$\ldots$]). The notations mentioned here are only related to the corresponding method. 

For FedProx, we set the parameter for proximal term $\mu$ to 0.001 (selecting from [0.1, 0.01, 0.001, 0.0001]). 

For Per-FedAvg, we set the step size $\alpha$ equal to the local learning rate. 

For FedRep, we set the number of local updates $\tau=5$ (selecting from [1, 2, 3, 4, 5, 6, 7, 8, 9, 10]) and set the step size $\alpha$ the same as the local learning rate. 

For pFedMe, we set its personalized learning rate to 0.01 (selecting from [0.1, 0.01, 0.001, 0.0001]), the additional parameter $\beta$ to 1.0 (selecting from [0.1, 0.5, 0.9, 1.0]), the regularization parameter $\lambda$ to 15 (selecting from [0, 0.1, 1, 5, 10, 15, 20, 50]) and the number of local computation $K$ to 5 (selecting from [1, 5, 10, 20]). 

For Ditto, we set the number of local epochs for personalized model $s=2$ (selecting from [1, 2, 3, 4, 5, 6, 7, 8, 9, 10]) and the coefficient of proximal term $\lambda=0.1$ (selecting from [0.0001, 0.001, 0.01, 0.1, 1, 10]). 

For FedAMP, we set the step size of gradient descent $\alpha_k$ to 1000 (selecting from [10000, 1000, 100, 10, 1, 0.1]), the regularization parameter $\lambda$ to 1 (selecting from [100, 10, 1, 0.1, 0.01]) and the parameter for attention-inducing function $\sigma$ to 0.1 (selecting from [1, 0.5, 0.1, 0.05, 0.01]). 

For FedPHP, we set the coefficient $\mu = 0.9$ (selecting from [0.0, 0.1, 0.2, 0.3, 0.4, 0.5, 0.6, 0.7, 0.8, 0.9, 1.0]) and $\lambda=0.01$ (selecting from [0.1, 0.05, 0.01, 0.005, 0.001]). 

For FedFomo, we set the number of received local models $M$ to the total number of clients. 

For APPLE, we set the loss scheduler type to `cos', directed relationship (DR) learning rate $\eta_2=0.01$ (selecting from [0.1, 0.05, 0.01, 0.005, 0.001]), $\mu=0.1$ (selecting from [1, 0.5, 0.1, 0.05, 0.01, 0.005, 0.001]), $L=0.2$ (selecting from [0.0, 0.1, 0.2, 0.3, 0.4, 0.5, 0.6, 0.7, 0.8, 0.9, 1.0]) and the number of received local models $M$ to the total number of clients. 

For PartialFed, we initialize the temperature parameter $\tau=5.0$ and anneal it to 0 with the decay rate of 0.965, based on the original paper. Besides, we set the updating frequency $f_m=8, f_s=2$ (selecting from [(9, 1), (8, 2), (7, 3), (6, 4), (5, 5), (4, 6), (3, 7), (2, 8), (1, 9)]) for it. 

For \method, we set the weights learning rate $\eta=1.0$ (selecting from [0.1, 1.0, 10.0]), random sample percent $s=80$ (selecting from [5, 10, 20, 40, 60, 80, 100]), \module range $p=1$ (selecting from [1, 2, ...]\footnote{The maximum search range varies using different backbones}).

\section{Dataset URLs}

MNIST\footnote{https://pytorch.org/vision/stable/datasets.html\#mnist}; Cifar10/100\footnote{https://pytorch.org/vision/stable/datasets.html\#cifarl}; Tiny-ImageNet\footnote{http://cs231n.stanford.edu/tiny-imagenet-200.zip}; AG News\footnote{https://pytorch.org/text/stable/datasets.html\#ag-news}. 

\section{Data Distribution Visualization}

\begin{figure*}[t]
	\centering
	\subfigure[$Dir(0.01)$]{\includegraphics[width=0.36\linewidth]{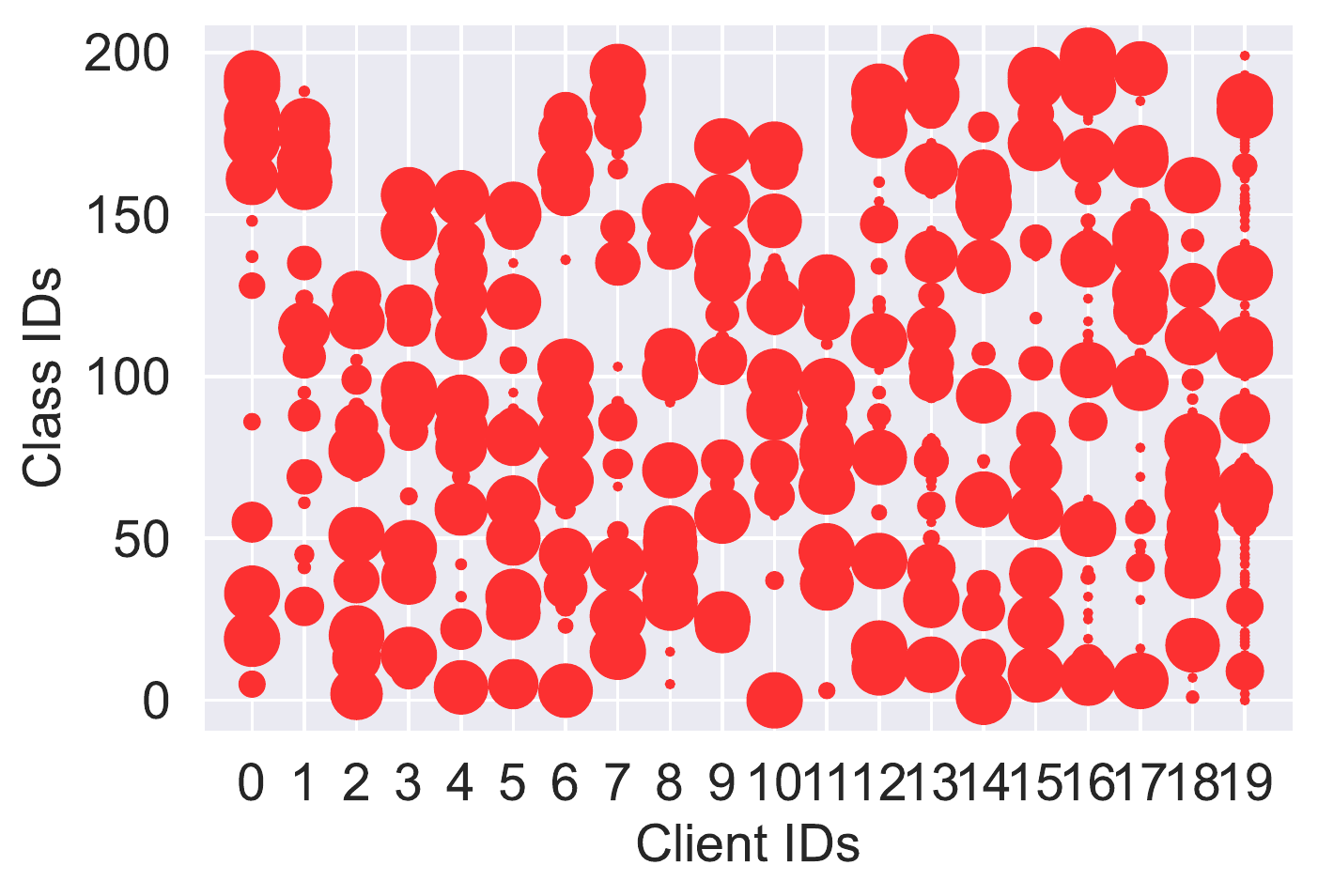}}
    \hfill
	\subfigure[$Dir(0.1)$]{\includegraphics[width=0.31\linewidth]{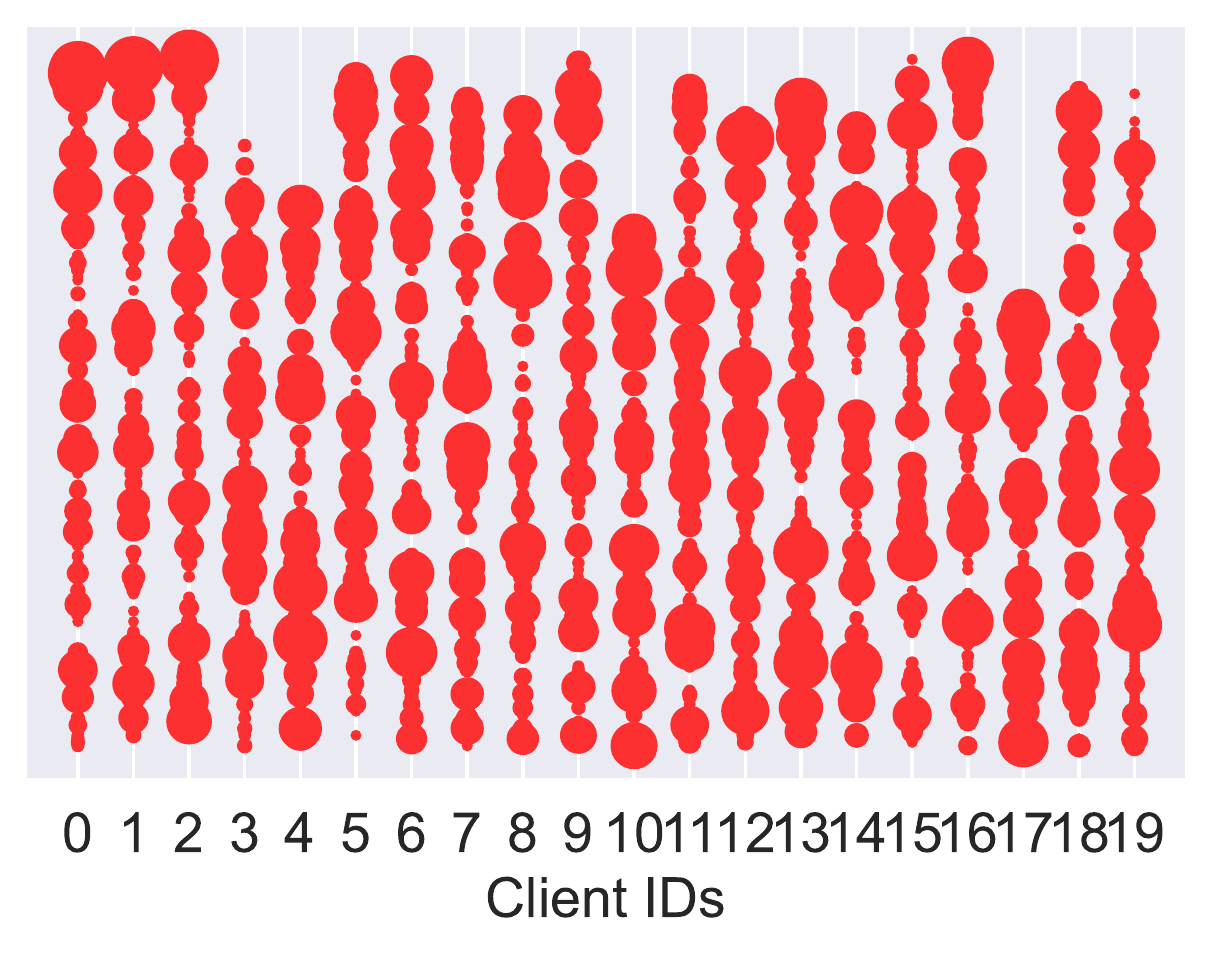}}
	\hfill
	\subfigure[$Dir(0.5)$]{\includegraphics[width=0.31\linewidth]{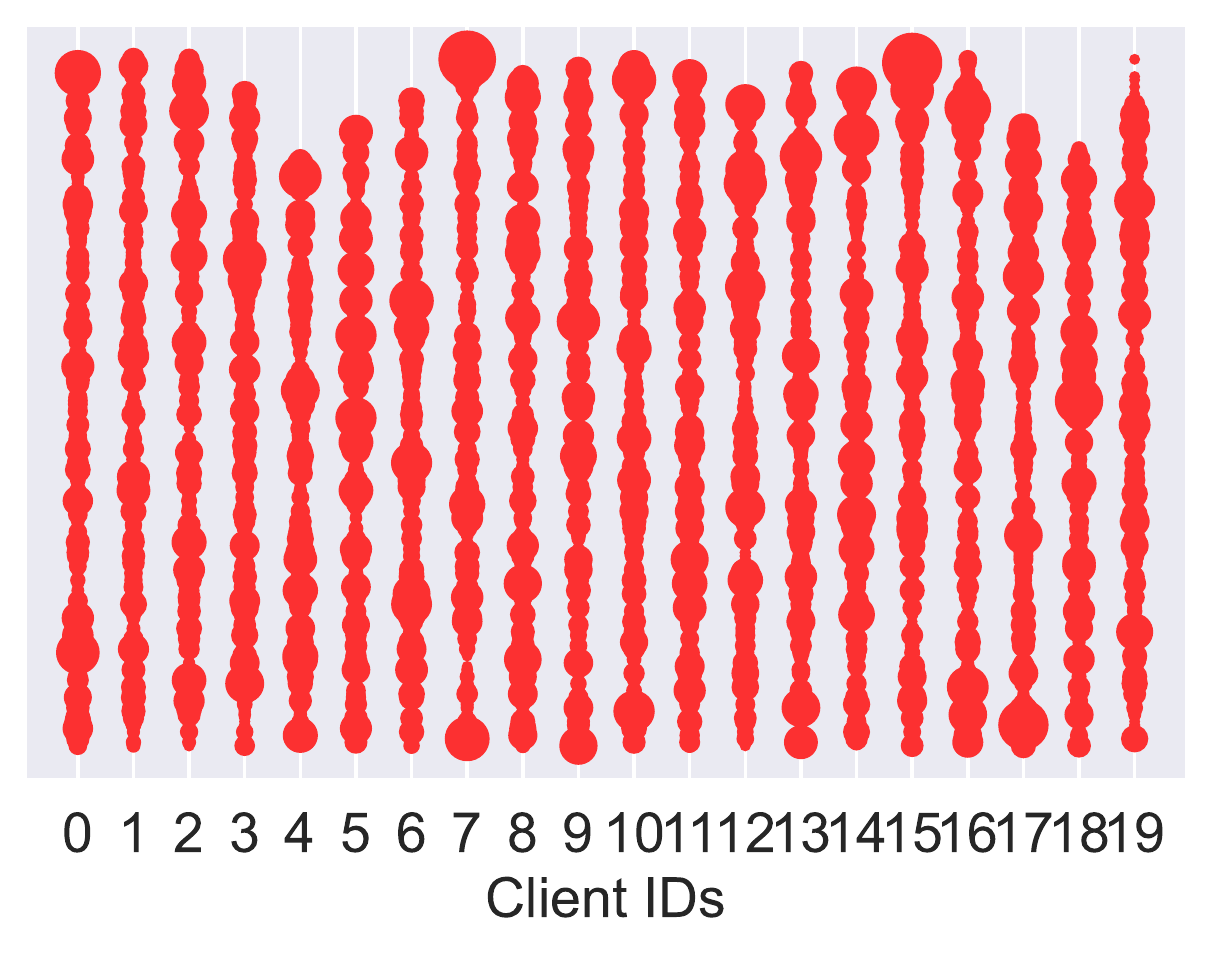}}
	
	\caption{The data distribution of each client on Tiny-ImageNet in three heterogeneous settings. The size of the circle represents the number of samples. With the $\beta$ in $Dir(\beta)$ increasing, the degree of heterogeneity decreases.}
	\label{fig:distribution}
\end{figure*}

\begin{figure*}[t]
	\centering
	\subfigure[MNIST]{\includegraphics[width=0.32\textwidth]{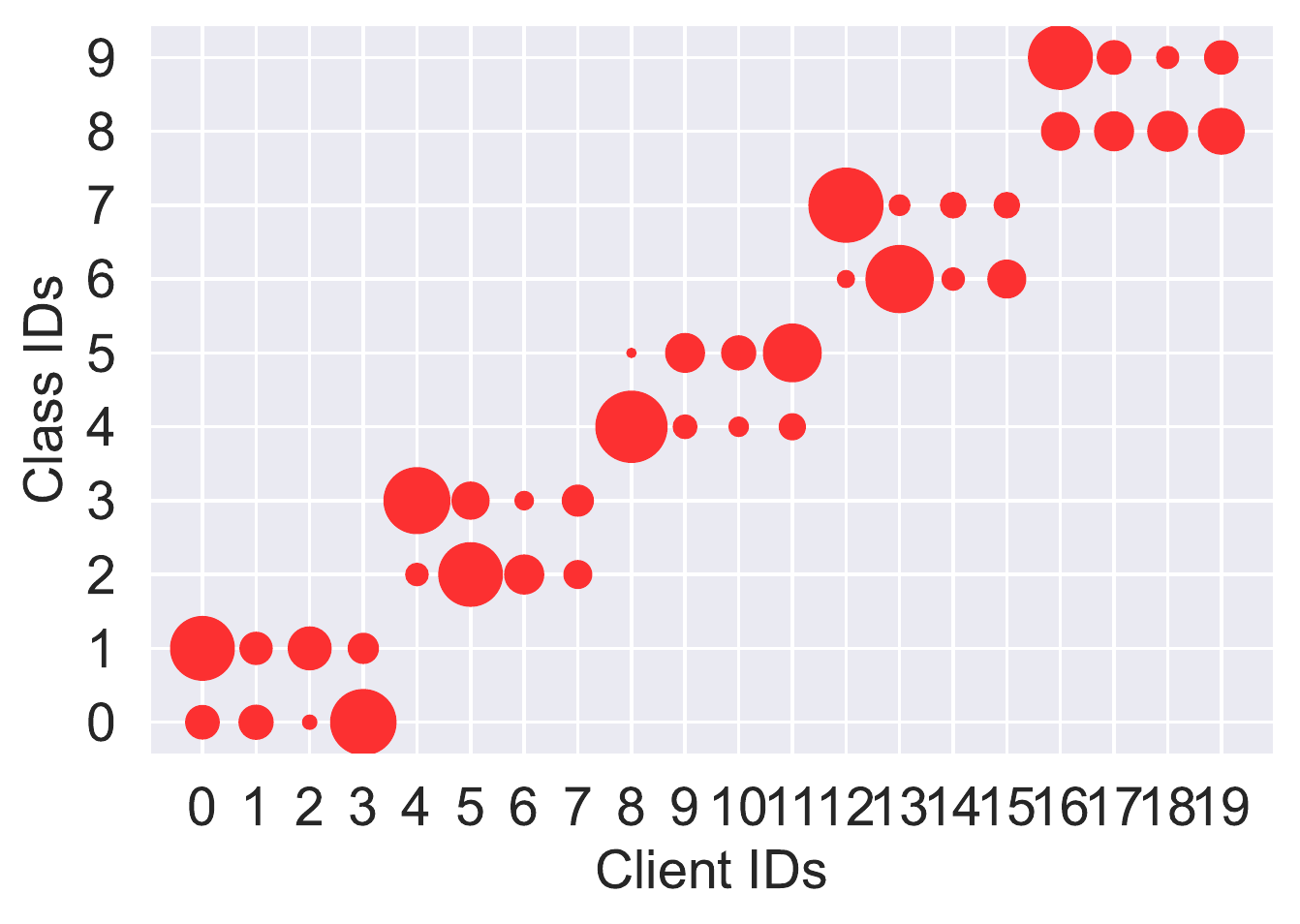}}
    \hfill
	\subfigure[Cifar10]{\includegraphics[width=0.32\textwidth]{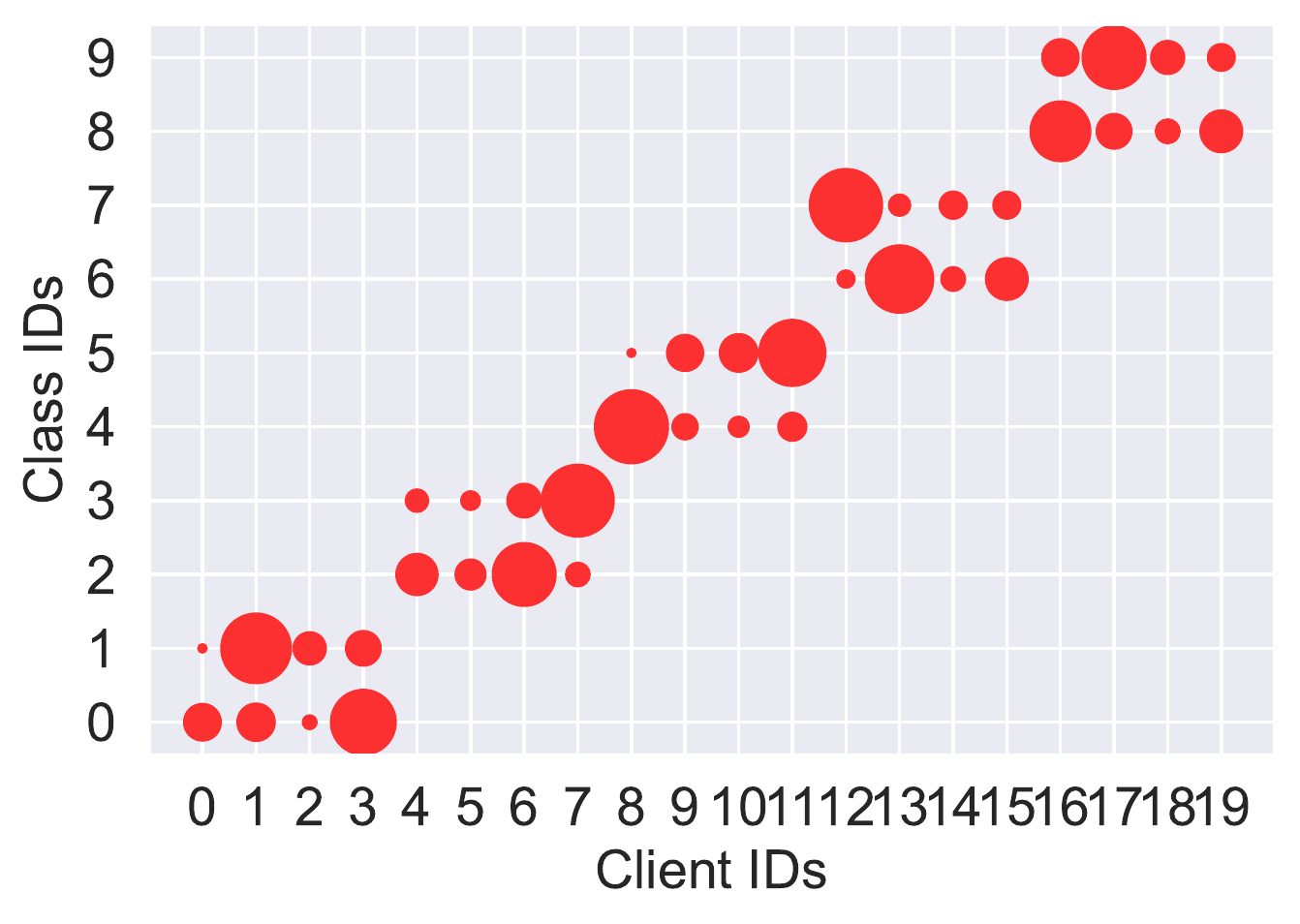}}
	\hfill
	\subfigure[Cifar100]{\includegraphics[width=0.32\textwidth]{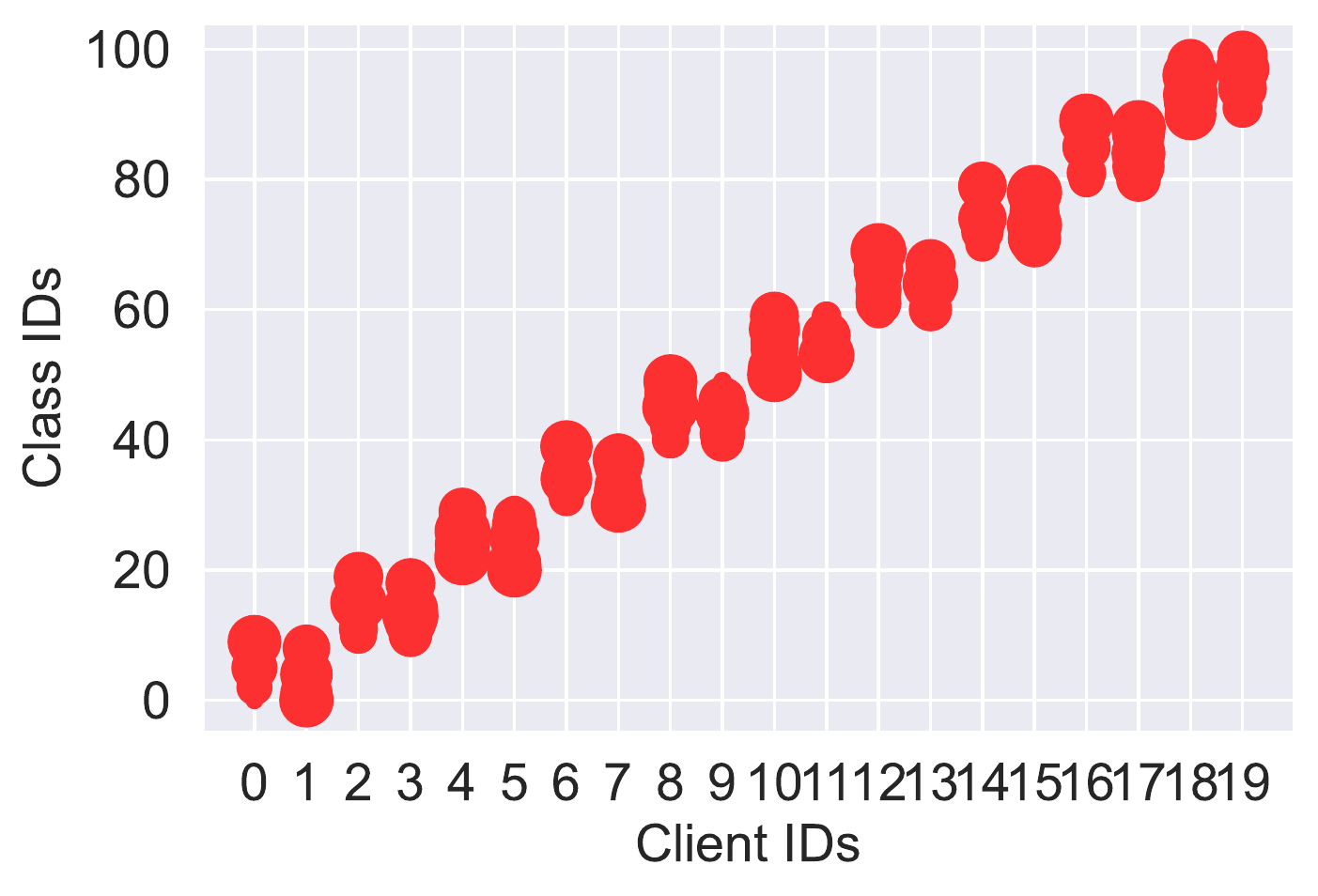}}
	\caption{The data distribution of each client on MNIST, Cifar10, and Cifar100 in pathological heterogeneous setting. The size of the circle represents the number of samples. }
	\label{fig:distribution-pathological}
\end{figure*}

\begin{figure*}[t]
	\centering
	\subfigure[MNIST]{\includegraphics[width=0.32\textwidth]{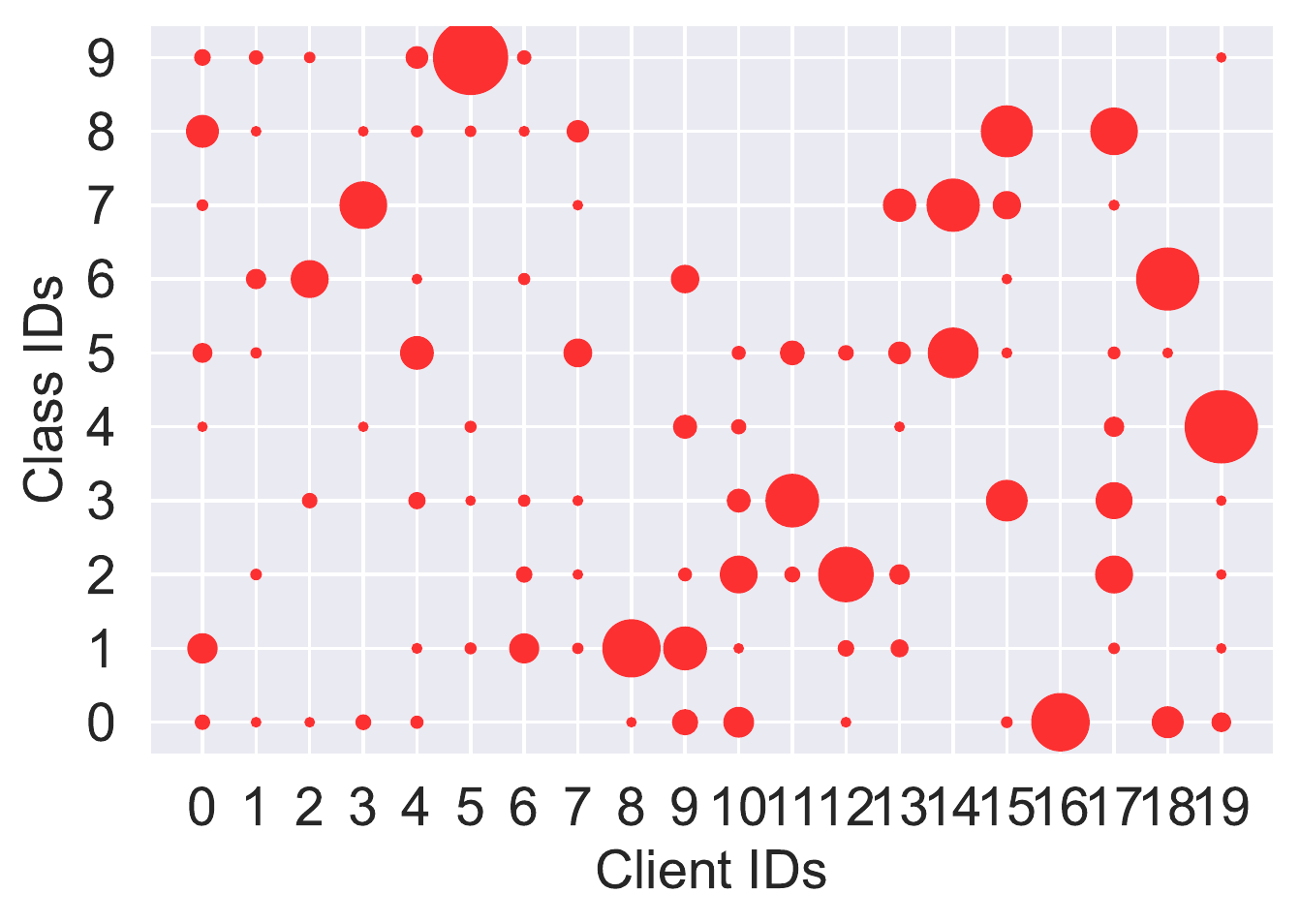}}
    \hfill
	\subfigure[Cifar10]{\includegraphics[width=0.32\textwidth]{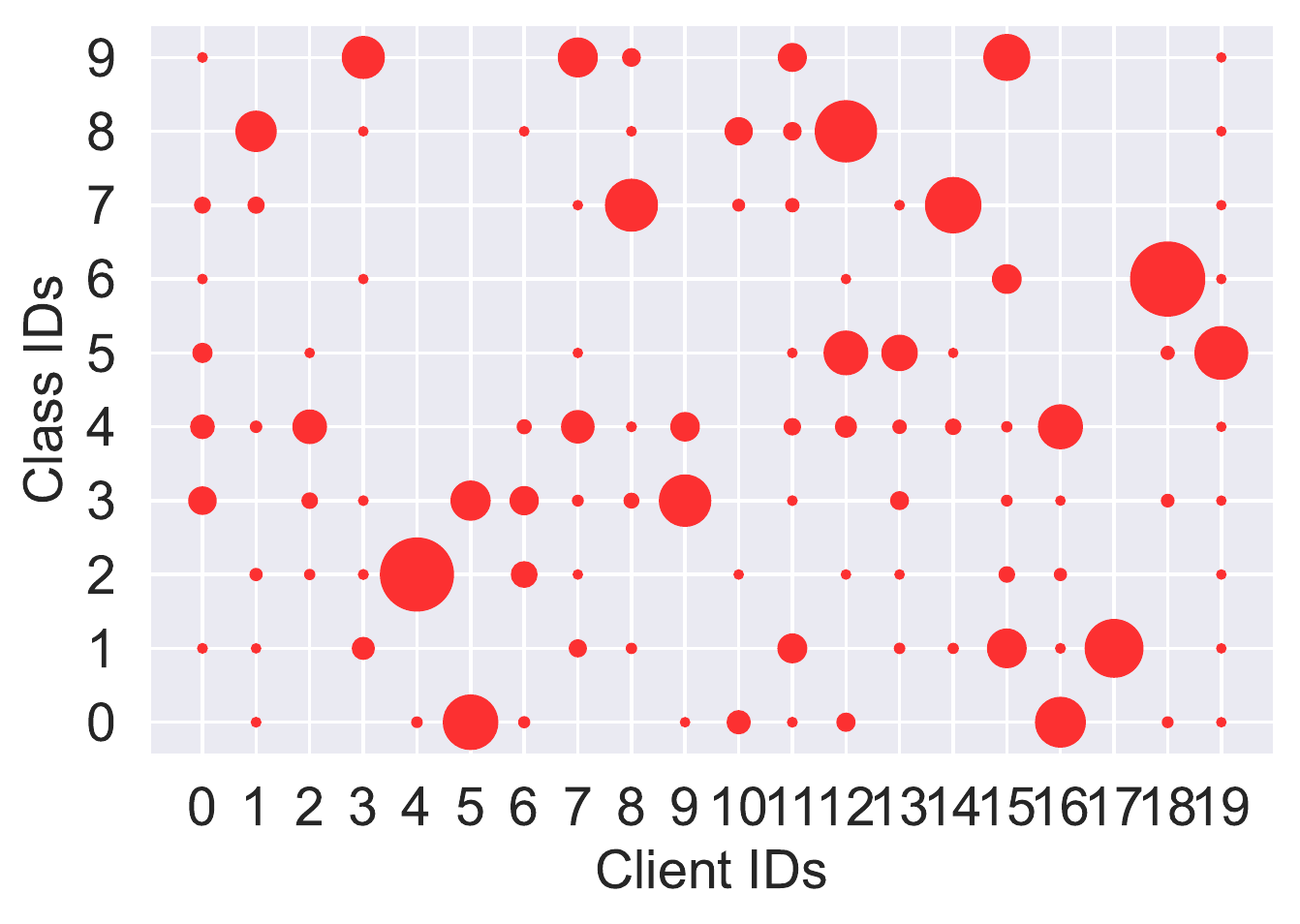}}
	\hfill
	\subfigure[Cifar100]{\includegraphics[width=0.32\textwidth]{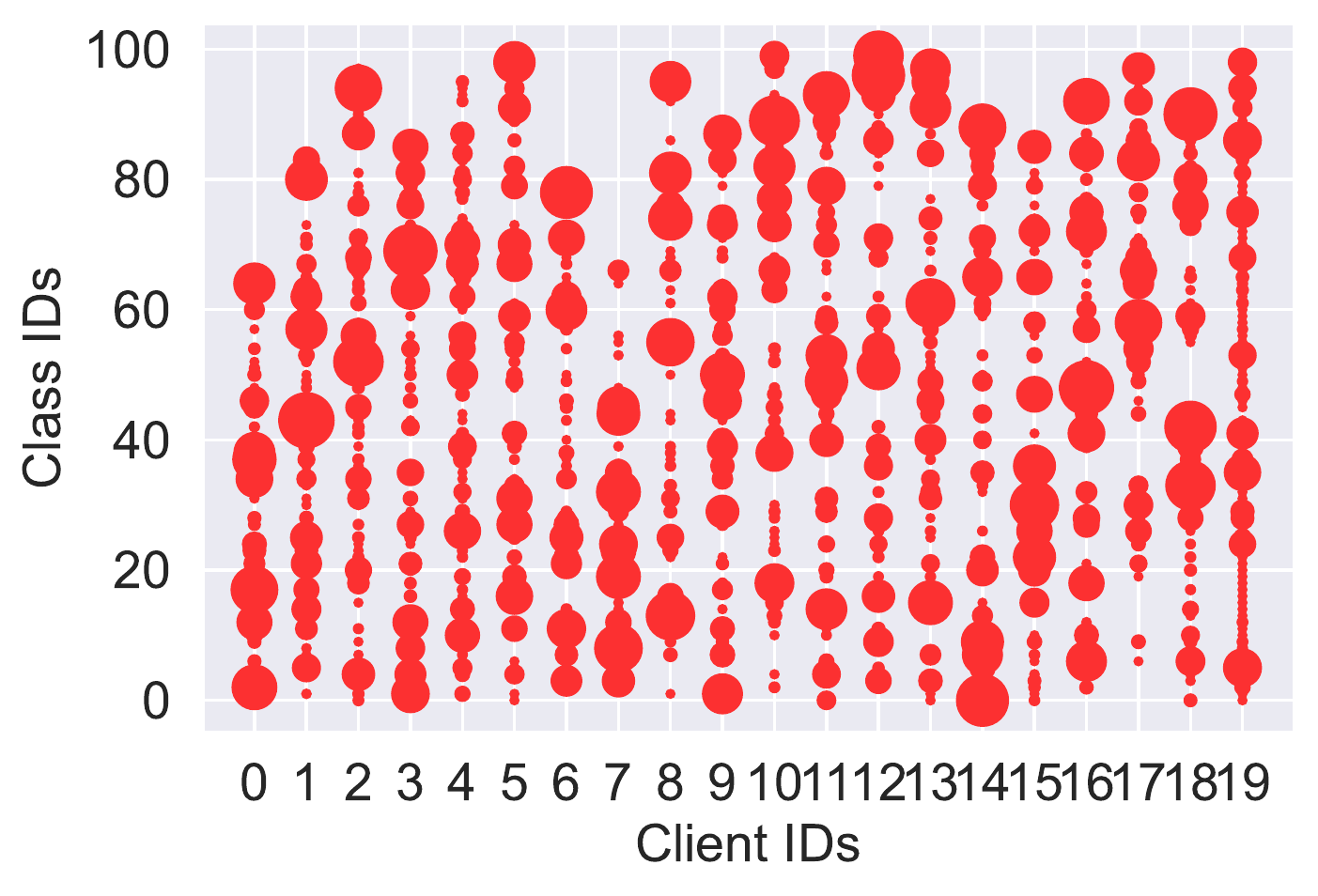}}
	\caption{The data distribution of each client on MNIST, Cifar10, and Cifar100 in default heterogeneous setting. The size of the circle represents the number of samples. }
	\label{fig:distribution-practical}
\end{figure*}

\begin{figure*}[t]
	\centering
	\subfigure[50 clients]{\includegraphics[width=\linewidth]{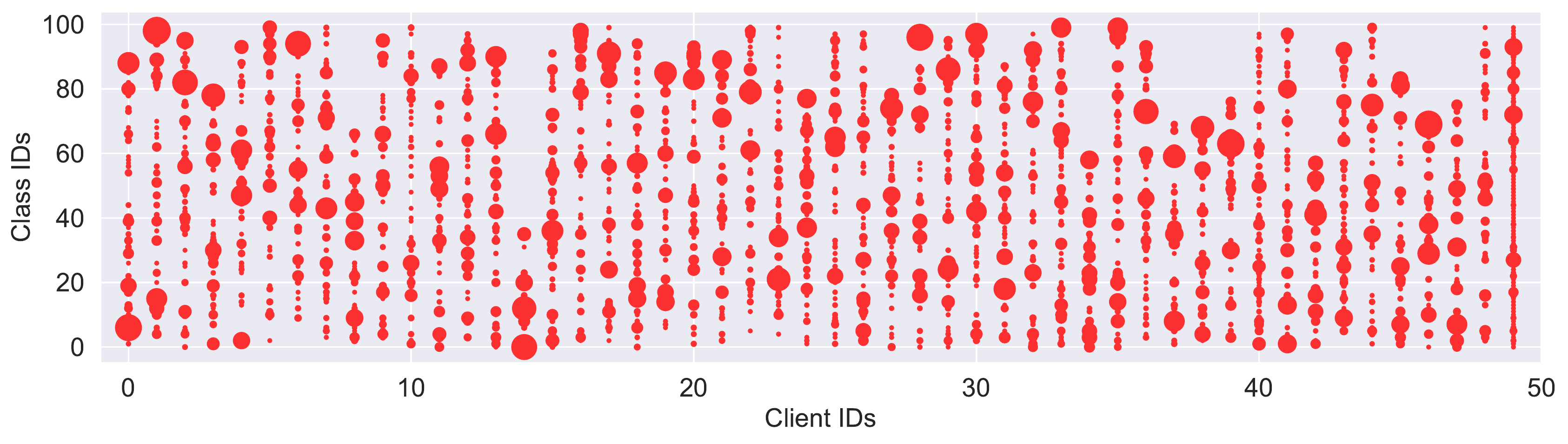}}
    \hfill
	\subfigure[100 clients]{\includegraphics[width=\linewidth]{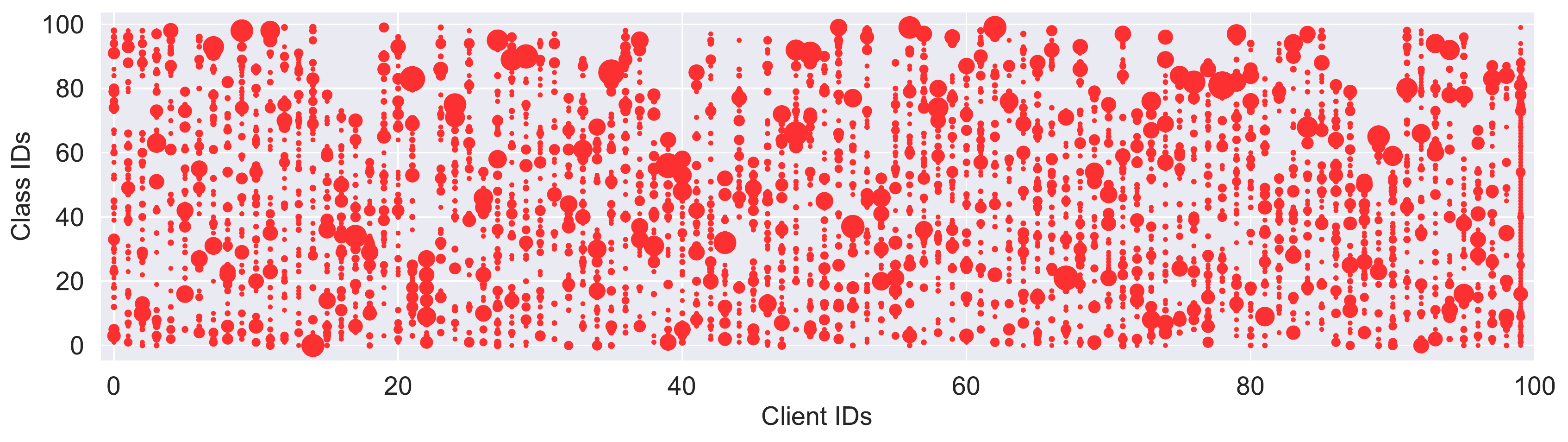}}
	\caption{The data distribution of each client on Cifar100 in default heterogeneous setting with 50 and 100 clients. The size of the circle represents the number of samples. }
	\label{fig:distribution-50100}
\end{figure*}

We illustrate the data distributions in the experiments in \Cref{fig:distribution}, \Cref{fig:distribution-pathological}, \Cref{fig:distribution-practical} and \Cref{fig:distribution-50100}. 

\end{document}